\documentclass[sigconf,natbib=true,anonymous=false,screen]{acmart}
\usepackage{xspace,balance,tabularx,multirow}
\usepackage{flushend}
\usepackage{tikz}
\usepackage{pgfplots}
\pgfplotsset{compat=1.16}
\usetikzlibrary{patterns}
\usepackage{subfig}
\usepackage[ruled, vlined, linesnumbered]{algorithm2e}
\usepackage{xcolor}
\usepackage{colortbl}
\usepackage{bbold}
\usepackage{enumitem}
\usepackage{tablefootnote}
\usepackage{upgreek,textgreek}
\usepackage{pifont}%
\usepackage[noabbrev]{cleveref}
\usepackage{titlecaps}
\usepackage{lipsum}

\pgfplotsset{every tick label/.append style={font=\tiny}}

\newlength{\starsize}
\newlength{\starspread}
\tikzset{starsize/.code={\setlength{\starsize}{#1}},
         starspread/.code={\setlength{\starspread}{#1}}}
\tikzset{starsize=1mm,
         starspread=3mm}
\pgfdeclarepatternformonly[\starspread,\starsize]%
  {my fivepointed stars}%
  {\pgfpointorigin}%
  {\pgfqpoint{\starspread}{\starspread}}%
  {\pgfqpoint{\starspread}{\starspread}}%
  {%
   \pgftransformshift{\pgfqpoint{\starsize}{\starsize}}
   \pgfpathmoveto{\pgfqpointpolar{18}{\starsize}}
   \pgfpathlineto{\pgfqpointpolar{162}{\starsize}}
   \pgfpathlineto{\pgfqpointpolar{306}{\starsize}}
   \pgfpathlineto{\pgfqpointpolar{90}{\starsize}}
   \pgfpathlineto{\pgfqpointpolar{234}{\starsize}}
   \pgfpathclose%
   \pgfusepath{fill}
  }

\newcommand{\eat}[1]{}

\newcommand{\argmax}[1]{\underset{#1}{\operatorname{arg}\,\operatorname{max}}\;}
\newcommand{\argmin}[1]{\underset{#1}{\operatorname{arg}\,\operatorname{min}}\;}

\makeatletter
\newcommand*\bigcdot{\mathpalette\bigcdot@{.5}}
\newcommand*\bigcdot@[2]{\mathbin{\vcenter{\hbox{\scalebox{#2}{$\m@th#1\bullet$}}}}}
\makeatother

\newcommand{\stitle}[1]{\vspace*{0.5em}\noindent{\underline{\bf #1.\/}}}

\newcommand{\eg}{{\it e.g.},\xspace}

\newcommand{\C}{\mathcal{C}\xspace}
\newcommand{\X}{\mathcal{X}\xspace}
\newcommand{\Y}{\mathcal{Y}\xspace}
\newcommand{\D}{\mathcal{D}\xspace}
\newcommand{\Hset}{\mathcal{H}\xspace}
\newcommand{\Sset}{\mathcal{S}\xspace}

\newcommand{\xvec}{\mathbf{x}\xspace}

\newcommand{\algo}{\textsf{NILC}\xspace}
\newcommand{\SAE}{\texttt{SAE}\xspace}
\newcommand{\DEC}{\texttt{DEC}\xspace}
\newcommand{\DCN}{\texttt{DCN}\xspace}
\newcommand{\CC}{\texttt{CC}\xspace}
\newcommand{\SCCL}{\texttt{SCCL}\xspace}
\newcommand{\USNID}{\texttt{USNID}\xspace}
\newcommand{\KCL}{\texttt{KCL}\xspace}
\newcommand{\MCL}{\texttt{MCL}\xspace}
\newcommand{\DTC}{\texttt{DTC}\xspace}
\newcommand{\GCD}{\texttt{GCD}\xspace}
\newcommand{\CDAC}{\texttt{CDAC+}\xspace}
\newcommand{\DeepAligned}{\texttt{DeepAligned}\xspace}
\newcommand{\SDC}{\texttt{SDC}\xspace}
\newcommand{\MTPCLNN}{\texttt{MTP-CLNN}\xspace}
\newcommand{\LatentEM}{\texttt{LatentEM}\xspace}
\newcommand{\LANID}{\texttt{LANID}\xspace}
\newcommand{\IntentGPT}{\texttt{IntentGPT}\xspace}

\newcommand{\SentenceBERT}{\texttt{SentenceBERT}\xspace}
\newcommand{\Instructor}{\texttt{Instructor}\xspace}

\newcommand{\GPTMini}{\texttt{GPT-4o-Mini}\xspace}
\newcommand{\GPT}{\texttt{GPT-4.1}\xspace}
\newcommand{\Gemini}{\texttt{Gemini-1.5-Pro}\xspace}
\newcommand{\Qwen}{\texttt{Qwen-Plus}\xspace}
\newcommand{\Deepseek}{\texttt{Deepseek-V3}\xspace}

\SetKwComment{tcp}{\(\triangleright\)\ }{}

\newenvironment{customlegend}[1][]{%
    \begingroup
    \csname pgfplots@init@cleared@structures\endcsname
    \pgfplotsset{#1}%
}{%
    \csname pgfplots@createlegend\endcsname
    \endgroup
}%

\def\addlegendimage{\csname pgfplots@addlegendimage\endcsname}

\makeatletter
\newcommand\footnoteref[1]{\protected@xdef\@thefnmark{\ref{#1}}\@footnotemark}
\makeatother

\let\oldnl\nl%
\newcommand{\nonl}{\renewcommand{\nl}{\let\nl\oldnl}}%

\SetKwComment{Comment}{/* }{ */}

\definecolor{myred}{HTML}{fd7f6f}
\definecolor{myred_new}{HTML}{D8D8D8}
\definecolor{myred_new2}{HTML}{D7191C}
\definecolor{myblue}{HTML}{7eb0d5}
\definecolor{mygreen}{HTML}{b2e061}
\definecolor{mypurple}{HTML}{bd7ebe}
\definecolor{myorange}{HTML}{ffb55a}
\definecolor{myyellow}{HTML}{ffee65}
\definecolor{mypurple2}{HTML}{beb9db}
\definecolor{mypink}{HTML}{fdcce5}
\definecolor{mycyan}{HTML}{8bd3c7}

\definecolor{myblue2}{HTML}{115f9a}
\definecolor{myred2}{HTML}{c23728}

\newcommand{\STAB}[1]{\begin{tabular}{@{}c@{}}#1\end{tabular}}

\AtBeginDocument{%
  \providecommand\BibTeX{{%
    \normalfont B\kern-0.5em{\scshape i\kern-0.25em b}\kern-0.8em\TeX}}}

\setcopyright{acmcopyright}
\copyrightyear{2018}
\acmYear{2018}
\acmDOI{XXXXXXX.XXXXXXX}

\acmPrice{15.00}
\acmISBN{978-1-4503-XXXX-X/18/06}

\acmSubmissionID{548}

\begin{document}

\title{\algo: Discovering New Intents with LLM-assisted Clustering}
\subtitle{Technical Report}

\author{Hongtao Wang}
\affiliation{%
  \institution{Hong Kong Baptist University}
  \country{Hong Kong SAR, China}
}
\email{cshtwang@comp.hkbu.edu.hk}
\orcid{0009-0002-1279-4357}

\author{Renchi Yang}
\authornote{Corresponding Author}
\affiliation{%
  \institution{Hong Kong Baptist University}
  \country{Hong Kong SAR, China}
}
\email{renchi@hkbu.edu.hk}
\orcid{0000-0002-7284-3096}

\author{Wenqing Lin}
\affiliation{%
  \institution{JD.com}
  \country{China}
}
\email{linwenqing.8@jd.com}
\orcid{0000-0003-4741-801X}

\settopmatter{printfolios=true}

\renewcommand{\shortauthors}{Wang et al.}

\begin{abstract}
{\em New intent discovery} (NID) seeks to recognize both new and known intents from unlabeled user utterances, which finds prevalent use in practical dialogue systems. Existing works towards NID mainly adopt a cascaded architecture, wherein the first stage focuses on encoding the utterances into informative text embeddings beforehand, while the latter is to group similar embeddings into clusters (i.e., intents), typically by $K$-Means. However, such a cascaded pipeline fails to leverage the feedback from both steps for mutual refinement,
and, meanwhile, the embedding-only clustering overlooks nuanced textual semantics, leading to suboptimal performance.

To bridge this gap, this paper proposes \algo{}, a novel clustering framework specially catered for effective NID.
Particularly, \algo{} follows an iterative workflow, in which clustering assignments are judiciously updated by carefully refining cluster centroids and text embeddings of uncertain utterances with the aid of {\em large language models} (LLMs).
Specifically, \algo{} first taps into LLMs to create additional {\em semantic centroids} for clusters, thereby enriching the contextual semantics of the Euclidean centroids of embeddings. Moreover, LLMs are then harnessed to augment hard samples (ambiguous or terse utterances) identified from clusters via rewriting for subsequent cluster correction.
Further, we inject supervision signals through non-trivial techniques {\em seeding} and {\em soft must links} for more accurate NID in the semi-supervised setting.
Extensive experiments comparing \algo{} against multiple recent baselines under both unsupervised and semi-supervised settings showcase that \algo{} can achieve significant performance improvements over six benchmark datasets of diverse domains consistently.
\end{abstract}

\ccsdesc[300]{Computing methodologies~Information extraction}
\ccsdesc[300]{Information systems~Query intent}
\ccsdesc[300]{Information systems~Clustering and classification}
\keywords{intent discovery, clustering, large language models}

\maketitle

\section{Introduction}
{\em New Intent Discovery} (NID)~\cite{mou2022generalized} 
aims to identify and group user utterances from conversational systems into both known and previously unseen intents (goals) in an open-world context, which outputs new intent labels or heuristics for faster annotations thereafter and automates the expensive manual annotation of tremendous data.
This task serves as building blocks to underpin a variety of applications, such as enhancing task-oriented dialogue systems (e.g., chatbots)~\cite{degand2020introduction} in e-commerce or other platforms~\cite{fan2025lanid,gao2021graphire,rodrigues2025intent},
refining search results through better query understanding~\cite{wang2023exploiting}, and elevating personalized services~\cite{li2021intention,li2022automatically} with better user modeling. 
The ability to dynamically discover and adapt to new intents is essential to 
ensure the effectiveness of these systems in an open-world environment where user needs are constantly evolving.

Existing solutions towards NID tasks can be broadly divided into two categories: {\em embedding-based methods} and {\em LLM-driven approaches}.
Embedding-based methods typically follow a cascaded pipeline.
They first encode all the utterances into a semantic space (i.e., text embeddings), followed by applying standard clustering algorithms like $K$-\texttt{Means} to group utterances into intent clusters~\cite{caron2018deep,zhang2022new,an2025unleashing}.
Most of them focus on pretraining or fine-tuning a text encoder, usually small language models, on available utterance corpora with limited or even without annotations, and thus, often struggle to capture the nuanced semantics hidden in utterances, particularly when dealing with technical or ambiguous utterances.
Moreover, while this modular methodology enjoys merits in simplicity and flexibility, it fails to leverage the results from two independent steps to optimize each other.

With the advent of {\em large language models} (LLMs), a promising way is to capitalize on the massive knowledge and superb comprehension ability of LLMs for NID, motivating a series of LLM-based methods~\cite{rodriguez2024intentgpt,de2023idas,lin2025spill,wang2025cequel}. 
However, as revealed in~\cite{lin2025spill,diaz2025k}, the use of text embeddings generated from LLMs in the cascaded pipeline for NID is not only computationally and financially expensive due to the sheer volume of utterance data in practice, but also suffers from poor performance caused by semantic drift as LLMs are trained on general-purpose corpora.
A recent work~\cite{rodriguez2024intentgpt} directly employs LLMs as classifiers to predict the intent categories of utterances, which often demands expensive fine-tuning for domain-specific understanding and appropriate granularity, and requires sophisticated prompting engineering with carefully curated examples to attain valid, stable, and decent results.
Very recently, several attempts~\cite{fan2025lanid,zou2025glean,liang2024synergizing} have been made towards combining the embedding-based and LLM-driven methodologies. 
Unfortunately, these efforts are limited by the static use of LLMs, primarily for embedding alignment, and the risk of inherited bias often outweighs the moderate performance gains.
In particular, most of the extant LLM-aided solutions are even inferior to the state-of-the-art embedding-based methods~\cite{zhang2023clustering}, as evidenced by our experiments.

In a nutshell, existing NID works are still suboptimal due to the problematic cascaded workflows, semantic ambiguity pervaded in user utterances, improper and ineffective use of LLMs.
This leads to a critical research question: can we devise a comprehensive framework that integrates embedding-based approaches with LLMs to overcome the limitations of both for enhanced NID cost-effectively?

\stitle{Our Contributions} In response to these challenges, we present \underline{N}ew \underline{I}ntent Discovery \underline{L}LM-assisted \underline{C}lustering (\algo{}), a novel and effective framework specialized for both unsupervised and semi-supervised NID.
Distinct from prior methods that merely focus on embedding learning or alignment, \algo{} mainly works on the clustering phase and operates in an iterative fashion, dynamically refining cluster assignments and text embeddings with the assistance of LLMs in each iteration, which enables the mutual optimization of both steps, and hence, circumvent the deficiency of cascaded architectures.
More specifically, 
in addition to the standard Euclidean centroids from embeddings, \algo{} introduces a {\em dual centroid scheme} that additionally generates a semantic centroid for each cluster by summarizing its theme with LLMs.
This design allows for capturing nuanced semantics neglected in embedding-only clustering, leading to more accurate cluster assignments.
On top of that, instead of using static text embeddings, \algo{} resorts to the {\em hard sample refinement} to identify hard samples (e.g., ambiguous utterances) based on current clustering results and leverage in-context learning (ICL) and generative capabilities of LLMs for subsequent augmentation and clustering. 
The cost-efficient use of LLMs for only clusters (exemplars therein) and hard samples not only 
bypasses the significant expense for all utterances, but also avoids introducing noise and contamination to high-quality samples that cause performance loss.
Furthermore, under the semi-supervised settings, \algo{} also innovatively incorporates supervision into the clustering stage through {\em seeding} and {\em soft must-link} constraints from labeled data, fostering improved NID performance.

Our extensive experiments on six benchmark datasets demonstrate that \algo{} remarkably outperforms a wide range of recent baselines based on embeddings or LLMs in both unsupervised and semi-supervised settings. The consistent superiority of our \algo{} across various domains highlights the effectiveness of our proposed framework, techniques, and optimizations.

\section{Related Work}

\subsection{Unsupervised Intent Discovery}
Unsupervised intent discovery partitions unlabeled utterances into intent categories. Early methods used traditional algorithms like $K$-\texttt{Means} \cite{mcqueen1967some} and {\em Agglomerative Clustering} (\texttt{AC}) \cite{gowda1978agglomerative} on static embeddings (\eg \texttt{TF-IDF} \cite{robertson1994some}, \texttt{GloVe} \cite{pennington2014glove}), but struggled with complex semantics.
To address this, deep clustering methods learn discriminative representations end-to-end, allowing the learned features to be tailored specifically for the NID task. For instance, \DEC \cite{xie2016unsupervised} uses an autoencoder for dimensionality reduction, \DCN \cite{shen2021semi} jointly performs feature learning and clustering, and \texttt{DeepCluster} \cite{caron2018deep} alternates between clustering to generate pseudo-labels and supervised training. However, these purely unsupervised methods cannot leverage prior knowledge from available labeled data, which is crucial for guiding the clustering process toward more semantically coherent intents and ensuring alignment with real-world application requirements.

\subsection{Semi-supervised Intent Discovery}
Incorporating labeled data, semi-supervised methods achieve superior performance through two main approaches: representation learning and clustering-based techniques.

Representation learning methods aim to learn a high-quality semantic space for text embeddings, often through contrastive learning. For example, \MTPCLNN \cite{zhang2022new} uses multi-task pre-training and a contrastive loss, while \USNID \cite{zhang2023clustering} employs a centroid-guided mechanism for self-supervised targets.
Clustering-based methods often use an iterative process. \CDAC \cite{lin2020discovering} uses pairwise constraints and a target distribution for refinement. \DeepAligned \cite{zhang2021discovering} iteratively refines cluster assignments after pre-training on labeled data. A key challenge is catastrophic forgetting, which \LatentEM \cite{zhou2023probabilistic} mitigates with a probabilistic EM framework treating intents as latent variables. \SDC \cite{an2025unleashing} leverages model bias from a pre-trained model to calibrate a trainable one. These methods, while powerful, primarily focus on representation learning and can struggle with ambiguous utterances.

\subsection{New Intent Discovery with LLMs}
The advent of LLMs has introduced a new paradigm for NID. These methods can be categorized based on how they utilize LLMs.
A key application is using LLMs as zero- or few-shot discoverers. For instance, \IntentGPT \cite{rodriguez2024intentgpt} employs a sophisticated prompting strategy with a few-shot sampler and feeds discovered intents back into the prompt for on-the-fly learning. However, this can be costly for large-scale applications.

A more common approach uses LLMs as knowledge distillers or supervisors to generate high-quality signals for training smaller models. For example, \texttt{IDAS} \cite{de2023idas} uses an LLM to generate abstractive summaries of utterances, cleaning the input for more effective clustering. \LANID \cite{fan2025lanid} queries an LLM for pairwise relationship labels to construct a contrastive fine-tuning task. Similarly, \texttt{GLEAN} \cite{zou2025glean} learns from diverse LLM feedback to refine model representations. While these hybrid methods are effective, they often lack a deep, iterative integration of the LLM's reasoning capabilities into the clustering process itself. Our work addresses this gap by using LLMs to iteratively strengthen ambiguous data points and enhance cluster assignments, offering greater effectiveness and interpretability.

\section{Preliminaries}
\subsection{Problem Statement}

\begin{figure}[!t]
    \centering
    \includegraphics[width=0.95\columnwidth]{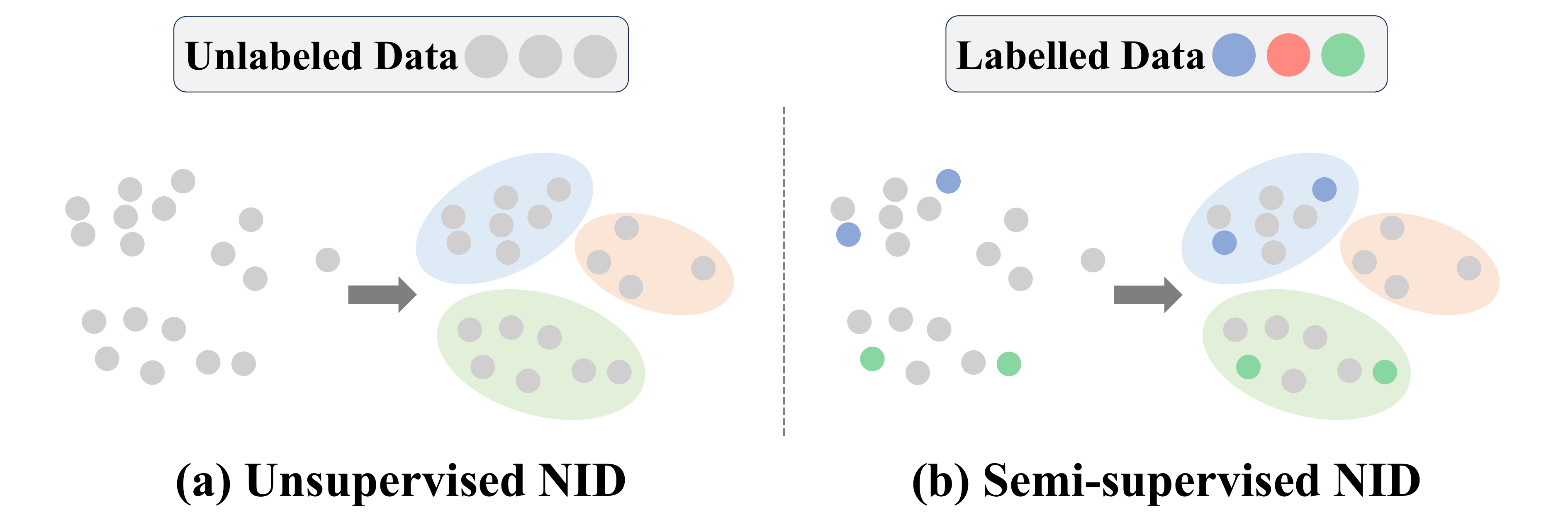} 
    \vspace{-2ex}
    \caption{Illustration of NID settings.}
    \label{fig:nid-problem}
\vspace{-3ex}
\end{figure}

Let $\Y_k$ be a set of $M$ known intent categories, and $\Y_u$ be a set of $K-M$ unknown intents. Accordingly, the set of all intents (also referred to as labels) is then denoted by $\Y=\Y_k\cup \Y_u$ and $|\Y|=K$. Given a set of labeled utterances from users $\D_{l}=\{(x_i,y_i)|y_i\in \Y_k\}_{i=1}^{|\D_{l}|}$, where $x_i$ is the $i$-th user utterance and $y_i$ stands for the corresponding intent, and a set $\D_{u}=\{x_i\}_{i=1}^{|\D_{u}|}$ of unlabeled user utterances, {\em New Intent Discovery} (NID) aims to identify all the intent categories $\Y$ (containing both known and novel intents) in $\D_{u}$. 

As depicted in Fig. \ref{fig:nid-problem}, NID has two settings, i.e., unsupervised and semi-supervised ones \cite{zhang2021discovering, lin2020discovering, song2023large}:

\stitle{Unsupervised NID} Under the unsupervised setting, there are no labeled utterances available, i.e., $\mathcal{D}_{l}=\emptyset$. The goal is to cluster $\mathcal{D}_{\text{test}}=\mathcal{D}_{u}$ into $K$ distinct intent categories.

\stitle{Semi-supervised NID}
In {semi-supervised} NID tasks, models are trained on training set $\mathcal{D}_{\text{train}}$ and evaluated on a {\em balanced} testing set $\mathcal{D}_{\text{test}}=\{(x_i,y_i)|y_i\in \Y\}_{i=1}^{|\mathcal{D}_{\text{test}}|}$,
where the training set is composed of both labeled and unlabeled utterances, i.e., $\mathcal{D}_{\text{train}} = \mathcal{D}_{l} \cup \mathcal{D}_{u}$. 
In particular, under this setting, the labeled data is limited, typically with a labeled ratio $\frac{|\mathcal{D}_{l}|}{|\mathcal{D}_{\text{train}}|}\le10\%$ and a {\em known class ratio} (KCR) $\frac{|\Y_k|}{K}\le 75\%$.
The trained models are expected to discriminate known intents from $\mathcal{D}_{\text{test}}$, while mining for new intents in the rest of utterances.

\begin{figure}[!t]
    \centering
    \includegraphics[width=0.96\columnwidth]{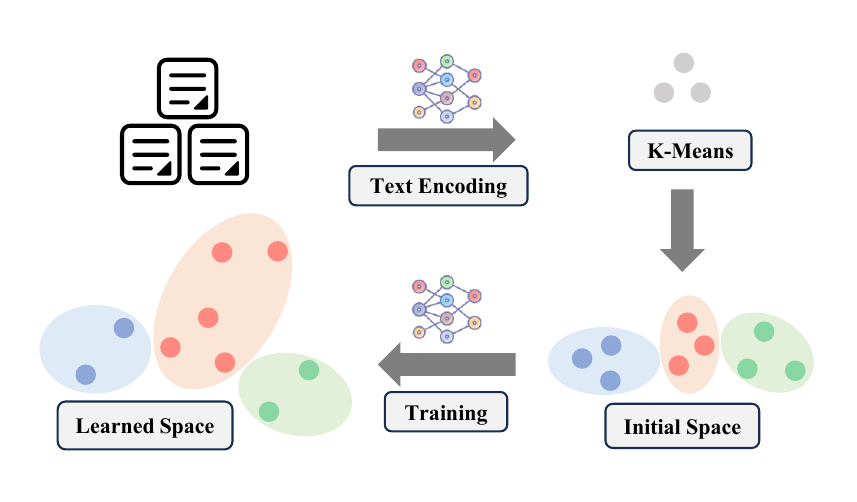} 
    \vspace{-5ex}
    \caption{Cascaded architecture for NID.}
    \label{fig:nid-pipeline1}
\vspace{-2ex}
\end{figure}

\subsection{Canonical Cascaded Architecture}

The majority of methods for NID follow a cascaded pipeline as illustrated in Fig.~\ref{fig:nid-pipeline1}, wherein the first stage focuses on mapping all utterances to a semantic embedding space, i.e., encoding them as text embeddings, while the second stage is a standard clustering algorithm (e.g., $K$-\texttt{Means}~\cite{mcqueen1967some}) applied to group these embeddings into intent categories. 
In the literature, the first phase is often referred to as {\em intent learning}.
A {\em pre-trained text encoder}, denoted as $\textsf{PTE}(\cdot)$, is trained or fine-tuned depending on the specific NID setting to produce the learned intent space.
In the unsupervised setting ($\mathcal{D}_{l}=\emptyset$), the encoder $\textsf{PTE}(\cdot)$ is trained on the unlabeled data $\mathcal{D}_{u}$. In the semi-supervised setting, it is trained on $\mathcal{D}_{\text{train}} = \mathcal{D}_{l} \cup \mathcal{D}_{u}$, leveraging information from both known intents ($\Y_k$) and unlabeled data. After training, the final encoder maps each utterance $x_i$ to a $d$-dimensional embedding vector $\xvec_i = \textsf{PTE}(x_i)$, yielding the learned intent space $\mathcal{X} = \{\xvec_i\}_{i=1}^N$. 

Representative NID works following the cascaded workflow include (i) \texttt{DCN}~\cite{shen2021semi} that jointly performs intent learning and clustering, (ii) \texttt{USNID}~\cite{zhang2023clustering} that designs a centroid-guided mechanism for self-supervised contrastive learning, and (iii) \texttt{LatentEM}~\cite{zhou2023probabilistic} that resorts to a probabilistic EM framework to optimize intent assignments. 
However, this cascaded architecture incurs inherent deficiencies. 
The decoupled nature of the intent learning and clustering stages prevents any feedback or mutual refinement. Consequently, the final performance is highly sensitive to the initial embedding quality and risks overlooking nuanced semantics, particularly for ambiguous or technical utterances.

\section{Methodology}

In this section, we present our clustering framework \algo for both unsupervised and semi-supervised NID. Firstly, we provide an overview of \algo{} in \S~\ref{sec:overview}, followed by elucidating algorithmic details for updating clusters and embeddings in \S~\ref{sec:centroids} and \ref{sec:refinement}, respectively. Subsequently, \S~\ref{sec:constraints} introduces our additional optimizations to inject semi-supervised signals into the clustering stage.
Finally, we conduct analyses regarding our clustering objective and the computational complexity of \algo{} in \S~\ref{sec:analysis}.

\begin{algorithm}[!t]
\small
\caption{\algo{} framework}\label{alg:nilc}
\KwIn{$\mathcal{D}_u \cup \mathcal{D}_l$, a pre-trained text encoder $\textsf{PTE}(\cdot)$, a large language model $\textsf{LLM}(\cdot)$, the number $K$ of intents, the number $T$ of iterations.}
\KwOut{$K$ intent clusters $\{\C_1,\C_2,\ldots,\C_K\}$.}
$\{\xvec_i\}_{i=1}^{N} \gets \textsf{PTE}(\mathcal{D}_u \cup \mathcal{D}_l)$  \tcp*{Encoding utterances}
$\{\C_k\}_{k=1}^K \gets K\text{-}\texttt{Means++}(\{\xvec_i\}_{i=1}^{N}, K)$ \tcp*{Initializing clusters}
\For{$t \gets 1$ to $T$}{
    \Comment{Dual Centroid Scheme (DCS)}
    \For{$k \gets 1$ to $K$}{
        Compute Euclidean centroid $\boldsymbol{\mu}_k$ \tcp*{Eq.~\eqref{eq:euclidean-centroid}}
        Generate cluster summary $s_k$ \tcp*{Eq.~\eqref{eq:clust-summary}}
        $\boldsymbol{\theta}_k \gets \textsf{PTE}(s_k)$ \tcp*{Eq.~\eqref{eq:sk-PTE}}
    }
    Update cluster assignments \tcp*{Eq.~\eqref{eq:clust-cost-i}}
    \Comment{Hard Sample Refinement (HSR)}
    Pick hard samples $\Hset$ with highest uncertainty in Eq.~\eqref{eq:uncertainty}\;
    \For{$\xvec_i$ in $\Hset$}{
        Generate refined utterance $\tilde{x}_i$ \tcp*{Eq.~\eqref{eq:refine-text}}
        $\tilde{\xvec}_i \gets \textsf{PTE}(\tilde{x}_i)$\;
        Conditionally update $\xvec_i$ to $\tilde{\xvec}_i$ \tcp*{Eq.~\eqref{eq:cond-update}}
    }
}
\end{algorithm}

\subsection{Synoptic Overview}\label{sec:overview}
\algo{} mainly focuses on the clustering of utterances for discovering new intents.
At a high level, \algo{} proceeds in an iterative fashion, where each iteration involves two main stages assisted with LLMs: (i) updating clusters, including cluster centroids and assignments, and (ii) refining text embeddings based on the current clustering results. 
Aside from the Euclidean centroids averaged over the text embeddings in clusters, the basic idea of the first stage is to introduce {\em semantic centroids} generated by LLMs to better summarize the main semantic themes of detected clusters for more accurate cluster assignments, which is referred to as {\em dual centroid scheme} (DCS).
The second stage is targeted to cope with {\em hard samples}, whose corresponding utterances are ambiguous or terse, rendering it hard to get informative text embeddings via the text encoder and determine their intent clusters subsequently with high confidence. The idea of \algo{} is to harness the extensive knowledge in LLMs to augment these utterances, and hence, obtain refined text embeddings for certain cluster assignments, dubbed as {\em hard sample refinement} (HSR). Notice that this HSR paradigm enables us to capitalize on the feedback from clustering to optimize the text embeddings, preventing the defects of traditional cascaded pipelines remarked earlier.

Algorithm~\ref{alg:nilc} summarizes the main steps in \algo{}. More concretely, given $\mathcal{D}_u \cup \mathcal{D}_l$, a $\textsf{PTE}(\cdot)$, an $\textsf{LLM}(\cdot)$, the numbers $K$ of intents and $T$ of iterations, \algo{} begins by encoding all the utterances $\{x_i\}_{i=1}^N$ in $\mathcal{D}_u \cup \mathcal{D}_l$ into text embeddings $\{\xvec_i\}_{i=1}^N$ (Line 1), and based thereon, initializes the $K$ clusters using $K$-\texttt{Means++} algorithm, a standard practice in clustering~\cite{arthur2007k} (Line 2).
After that, \algo{} starts an iterative procedure for updating clusters using DCS and refining hard samples using HRS (Lines 3-13).
In each of the $T$ iterations, the DCS phase will re-calculate the Euclidean centroids $\{\boldsymbol{\mu}_k\}_{k=1}^K$ of the current clusters $\{\C_k\}_{k=1}^K$, and generate semantic centroids $\{\boldsymbol{\theta}_k\}_{k=1}^K$ through $\textsf{LLM}(\cdot)$ and $\textsf{PTE}(\cdot)$ (Lines 4-7).
The cluster assignments of all utterance samples will later be updated accordingly at Line 8.
Subsequently, \algo{} proceeds to the HSR stage, in which a small set of hard samples $\Hset$ is first identified, followed by a refinement of their textual contents leveraging $\textsf{LLM}(\cdot)$ and $\textsf{PTE}(\cdot)$, as well as a conditional update of their text embeddings (Lines 9-13).

In the succeeding subsections, we elaborate on the details of generating semantic centroids with LLMs, updating cluster assignments, as well as identifying and refining hard samples with LLMs.

\begin{figure}[!t]
    \centering
    \includegraphics[width=0.99\columnwidth]{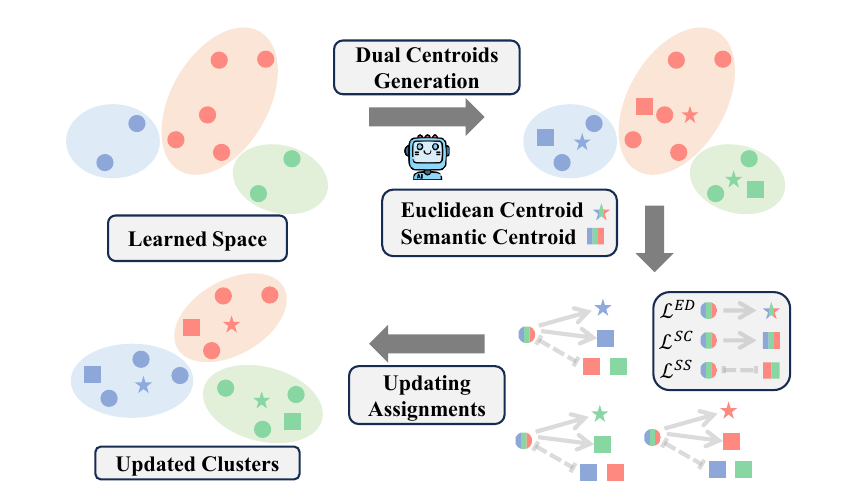} 
    \vspace{-2ex}
    \caption{Dual Centroid Scheme in \algo.}
    \label{fig:nid-pipeline2}
\vspace{-2ex}
\end{figure}

\subsection{Cluster Update with Dual Centroids}\label{sec:centroids}

As displayed in Fig.~\ref{fig:nid-pipeline2}, the DCS is to derive an Euclidean centroid $\boldsymbol{\mu}_k$ and a semantic centroid $\boldsymbol{\theta}_k$ for each of the $K$ intent clusters $\{\C_1,\C_2,\ldots,\C_K\}$. Akin to $K$-\texttt{Means}, the Euclidean centroid $\boldsymbol{\mu}_k$ of cluster $\C_k$ is computed as the mean of the embedding vectors of the samples therein, i.e., 
\begin{equation}\label{eq:euclidean-centroid}
\boldsymbol{\mu}_k = \frac{1}{|\C_k|} \sum_{x_i \in \C_k} \xvec_i.
\end{equation}
Although the text embeddings $\{\xvec_i\}_{i=1}^N$ are often obtained via the domain- or task-specific \textsf{PTE}, directly averaging them as centroids will obscure nuanced textual relationships. Moreover, such centroids have no explicit and precise semantic themes~\cite{diaz2025k}, i.e., the underlying intent categories are indefinite, rendering some utterances likely to be misassigned to the intent clusters.
As a remedy, we directly summarize the utterances in each cluster $\C_k$ as a semantic centroid $\boldsymbol{\theta}_k$ to complement the Euclidean centroid $\boldsymbol{\mu}_k$.

\begin{figure}[!t]
    \centering
    \includegraphics[width=0.99\columnwidth]{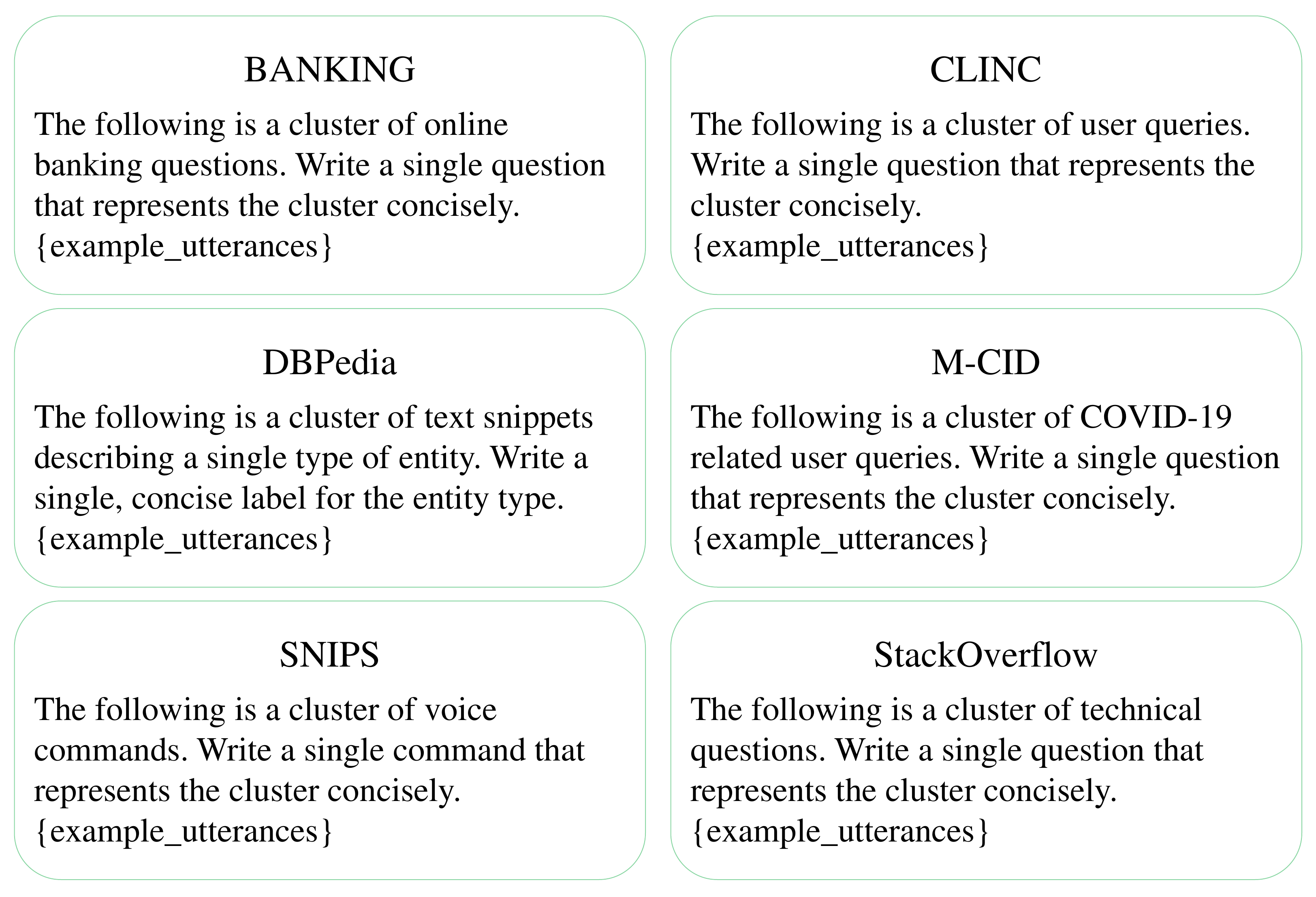} 
    \vspace{-2ex}
    \caption{Prompts for generating cluster summaries.}
    \label{fig:nid-summary-template}
\vspace{-1ex}
\end{figure}

\stitle{Generating Semantic Centroids with LLMs} 
Instead of feeding all the utterances in each cluster to LLMs, which are both financially and computationally expensive, \algo{} cherry-picks a subset $\Sset_k\subset \C_k$ of $|\Sset_k|$ (typically $|\Sset_k| = 10$) exemplars for cluster $\C_k$ based on a certain selection strategy\footnote{The details and evaluations are deferred to Appendix~\ref{sec:selection-strategy}, 
\ref{sec:case-study-evolution}, and 
\ref{sec:empiracal-study-mapping-sampling}.}, such as $K$-\texttt{Means} for maximal diversity, {\em maximal marginal Relevance} (MMR) for balancing relevance and diversity, {\em mean average distance} (MAD), and {\em nearest neighbors} (NN), etc. As exemplified in Fig.~\ref{fig:nid-summary-template}, together with a summary generation prompt $p_{\text{smry}}$, these samples are subsequently forwarded to an LLM to generate a textual summary $s_k$ for cluster $\C_k$:
\begin{equation}\label{eq:clust-summary}
s_k = \textsf{LLM}(p_{\text{smry}}, \Sset_k).
\end{equation}
Intuitively, by unleashing the extensive knowledge and remarkable summarization capacity of LLMs, $s_k$ can convey the intent semantics of cluster $C_k$ more precisely. We then encode it as the semantic centroid in the form of embedding vectors by the \textsf{PTE}:
\begin{equation}\label{eq:sk-PTE}
\boldsymbol{\theta}_k = \textsf{PTE}(s_k),
\end{equation}
which has the potential to assist us in correcting the misassignments of utterance samples in cluster $\C_k$.

\stitle{Updating Cluster Assignments}
Given dual centroids $\boldsymbol{\mu}_k$ and $\boldsymbol{\theta}_k$ for each cluster $\C_k$, the next task is to update the cluster assignments for all utterance samples. The assignments are determined through minimizing a joint clustering cost function based on dual centroids, namely,
\begin{equation}\label{eq:clust-cost-i}
y_i = \argmin{1\le k\le K}{ f(\xvec_i)= \mathcal{L}^{\text{ED}}_i + \alpha \cdot \mathcal{L}^{\text{SC}}_i + \beta \cdot \mathcal{L}^{\text{SS}}_i}
\end{equation}
where $\alpha$ and $\beta$ stand for coefficients adjusting the importance of the terms $\mathcal{L}^{\text{SC}}_i$ and $\mathcal{L}^{\text{SS}}_i$, respectively. Specifically, the first term $\mathcal{L}^{\text{ED}}_i$ measures the Euclidean distance of sample $\xvec_i$ to any Euclidean centroid $\boldsymbol{\mu}_k$, i.e.,
\begin{equation}
    \mathcal{L}^{\text{ED}}_i = \left\|\xvec_i - \boldsymbol{\mu}_k\right\|^2.
\end{equation}

The other two terms, $\mathcal{L}^{\text{SC}}_i$ and $\mathcal{L}^{\text{SS}}_i$, quantify the semantic dissimilarities between sample $\xvec_i$ and the cluster from the intra- and inter-cluster perspectives, respectively.
More specifically, $\mathcal{L}^{\text{SC}}_i$ is defined as the dissimilarity of utterance $\xvec_i$ and the semantic centroid $\boldsymbol{\theta}_k$ of cluster $\C_k$:
\begin{equation}
    \mathcal{L}^{\text{SC}}_i = 1 - \textsf{cos}\left(\xvec_i, \boldsymbol{\theta}_{k}\right).
\end{equation}
Intuitively, this term is minimized when we assign the cluster with the highest semantic similarity, i.e., most relevant theme, to $\xvec_i$, which in turn encourages increasing the intra-cluster semantic cohesion of resulting clusters.

Conversely, the semantic separation term $\mathcal{L}^{\text{SS}}_i$ serves as a repulsive force to enhance inter-cluster distinctiveness, which is formulated as
\begin{equation}
    \mathcal{L}^{\text{SS}}_i = \textsf{cos}(\xvec_i, \boldsymbol{\theta}_{\text{nbr}(k)}),
\end{equation}
where $\boldsymbol{\theta}_{\text{nbr}(k)}$ represents the other semantic centroid that is closest to the semantic centroid $\boldsymbol{\theta}_k$ of the cluster assigned to $\xvec_i$. 
Through additionally minimizing the similarity of sample $\xvec_i$ to the nearest neighboring semantic centroid of $\boldsymbol{\theta}_k$, \algo{} ensures all utterances within cluster $\C_k$ radically differ from the rest of the intent clusters in terms of semantic themes.

\begin{figure}[!t]
    \centering
    \includegraphics[width=0.99\columnwidth]{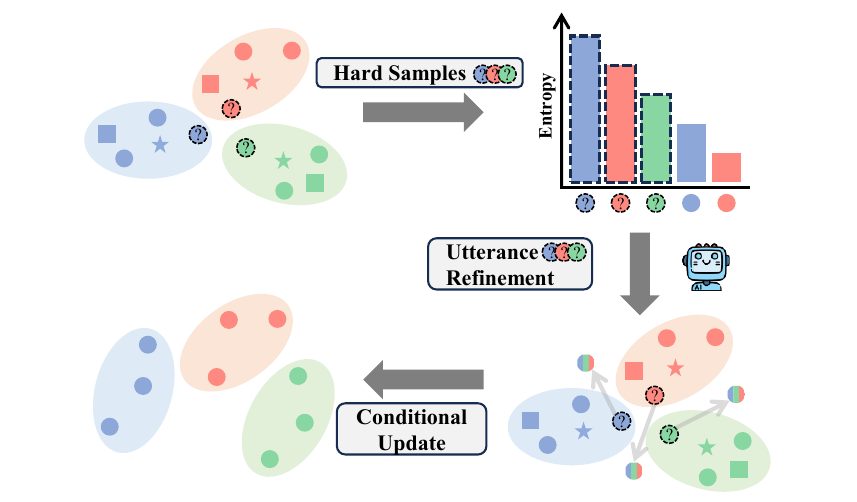} 
    \vspace{-3ex}
    \caption{Hard Sample Refinement in \algo.}
    \label{fig:nid-pipeline3}
\vspace{-2ex}
\end{figure}

\subsection{Hard Sample Refinement with LLMs}\label{sec:refinement}

In updated clusters, there usually exist a number of {\em hard samples}, which are located on the boundary of clusters, as their corresponding input utterances are usually noisy, ambiguous, overly sketchy, abbreviation-heavy, or cryptic due to the presence of jargon and slang, etc. 
For instance, the utterance ``Confusion regarding laziness'' from StackOverflow is ambiguous, as ``laziness'' could refer to general performance issues or a specific concept in functional programming. This ambiguity can lead to misclassification into a related but incorrect cluster (e.g., LINQ queries) \footnote{Case studies are provided in Appendix~\ref{sec:case-study-hsr}.}. 
As such, these samples are likely to cause the ``centroid shift'' since Euclidean centroids are simply averaged over within-cluster text embeddings, and thus, slightly distort the themes of the intent clusters and result in erroneous assignments.
As depicted in Fig.~\ref{fig:nid-pipeline3}, HSR mitigates this issue through a refinement scheme consisting of the following three steps. With HSR, the above utterance example can be rewritten as a more precise one ``Understanding laziness in functional programming languages like Haskell,'' making it more likely to be correctly classified.

\stitle{Identifying Hard Samples}
To identify hard utterance samples, we propose to assess the uncertainty of each sample's clustering assignment and select a subset $\Hset$ comprising the top-$\delta$ (typically $\delta = 10$) samples with the highest uncertainty. Let $P(\C_k|\xvec_i)$ be the posterior probability of assigning sample $\xvec_i$ to intent cluster $\C_k$, i.e., soft assignment. Following the practice in {\em deep clustering}~\cite{xie2016unsupervised}, we employ the classic Gaussian kernel to measure the similarity between sample $\xvec_i$ and centroid $\boldsymbol{\mu}_k$, and then transform it into the posterior probability via the softmax function:
\[
P(\C_k|\xvec_i) = \frac{\exp(-\|\xvec_i - \boldsymbol{\mu}_k\|^2)}{\sum_{\ell=1}^K \exp(-\|\xvec_i - \boldsymbol{\mu}_\ell\|^2)}.
\]
A simple and effective way to measure the assignment uncertainty is the prominent {\em Shannon entropy}~\cite{shannon1948mathematical}. In mathematical terms, for each sample $\xvec_i$, the uncertainty is formulated by
\begin{equation}\label{eq:uncertainty}
H(\xvec_i) = -\sum_{k=1}^K P(\C_k|\xvec_i)\cdot \log P(\C_k|\xvec_i).
\end{equation}
A high entropy value indicates that the assignment probabilities are evenly distributed among all $K$ clusters, implying high uncertainty. Intuitively, the larger $H(\xvec_i)$ is, the less confident we are about the cluster assignment for utterance $x_i$.

\stitle{Context-aware Rewriting}
After the obtainment of the hard sample set $\Hset$, \algo{} enters into rewriting their original utterances with LLMs. Inspired by a recent finding that editing input text can be more effective than generating it from scratch~\cite{ye2023enhancing}, we introduce a context-aware rewriting mechanism that structures the LLM's task to emulate a cognitive process of analysis before synthesis. This ``judge-then-rewrite'' paradigm first compels the model to analytically determine the most plausible cluster for a given hard sample before reformulating the utterance, thereby enhancing the relevance and quality of the generated text.

\begin{figure}[!t]
    \centering
    \includegraphics[width=0.97\columnwidth]{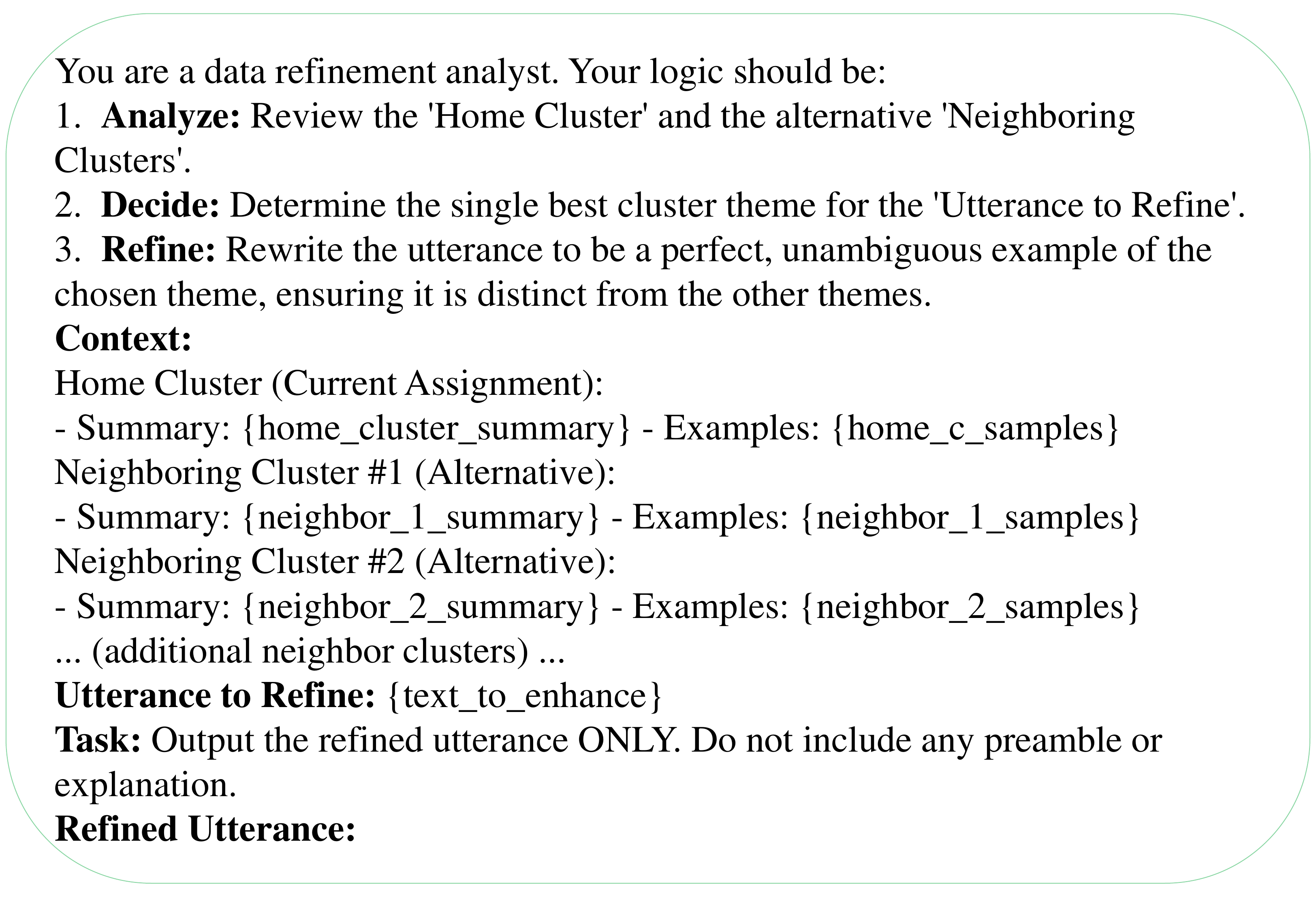} 
    \vspace{-2ex}
    \caption{Prompt template for HSR.}
    \label{fig:nid-refine-template}
\vspace{-2ex}
\end{figure}

This process is guided by a meticulously designed prompt template, denoted as $p_{\text{ref}}$ and detailed in Fig.~\ref{fig:nid-refine-template}. For each hard sample (i.e., ambiguous utterance) $x_i$, the prompt is furnished with rich in-context information pertinent to its currently assigned cluster $\C_k$, referred to as ``home'' cluster, and $K_{\text{nbr}}$ nearest neighboring clusters $\{\C_j\}_{j=1}^{K_{\text{nbr}}}$ (typically $K_{\text{nbr}}=10$), dubbed as {\em neighboring clusters}. More precisely, we include the semantic summary $s_k$ and a set of representative exemplars $\Sset_k$ of home cluster $\C_k$ into prompt $p_{\text{ref}}$, as well as its counterparts $\{s_j, \Sset_j\}_{j=1}^{K_{\text{nbr}}}$ for the $K_{\text{nbr}}$ neighboring clusters.
Afterwards, the LLM is invoked to generate the refined utterance $\tilde{x}_i$ by
\begin{equation}\label{eq:refine-text}
\tilde{x}_i = \textsf{LLM}\left(p_{\text{ref}}, x_i\right).
\end{equation}

\stitle{Conditional Update} Lastly, the revised version $\tilde{x}_i$ of utterance $x_i$ is encoded by $\textsf{PTE}(\cdot)$ previously used as the new text embedding $\tilde{\xvec}_i$. Instead of blindly updating $\xvec_i$ to $\tilde{\xvec}_i$ that might introduce noise from LLMs, we substitute $\tilde{\xvec}_i$ for original $\xvec_i$ for subsequent clustering steps only if $\tilde{\xvec}_i$ can leads to a reduction in the clustering cost $f(\xvec_i)$ of $x_i$ defined in Eq.~\eqref{eq:clust-cost-i}. Mathematically,
\begin{equation}\label{eq:cond-update}
\xvec_i =
\begin{cases} 
\tilde{\xvec}_i & \text{if } \min_{1\le k\le K} f(\tilde{\xvec}_i) < \min_{1\le k\le K} f({\xvec}_i), \\
\xvec_i & \text{otherwise.}
\end{cases}
\end{equation}
This conditional update enforces that the clustering cost for each sample is greedily minimized even when we involve LLMs for data augmentation, i.e., rewriting utterances, thereby ensuring stable and reliable clustering results.

\subsection{Optimizations for Semi-Supervised NID}\label{sec:constraints}

\begin{figure}[!t]
    \centering
    \includegraphics[width=0.5\textwidth]{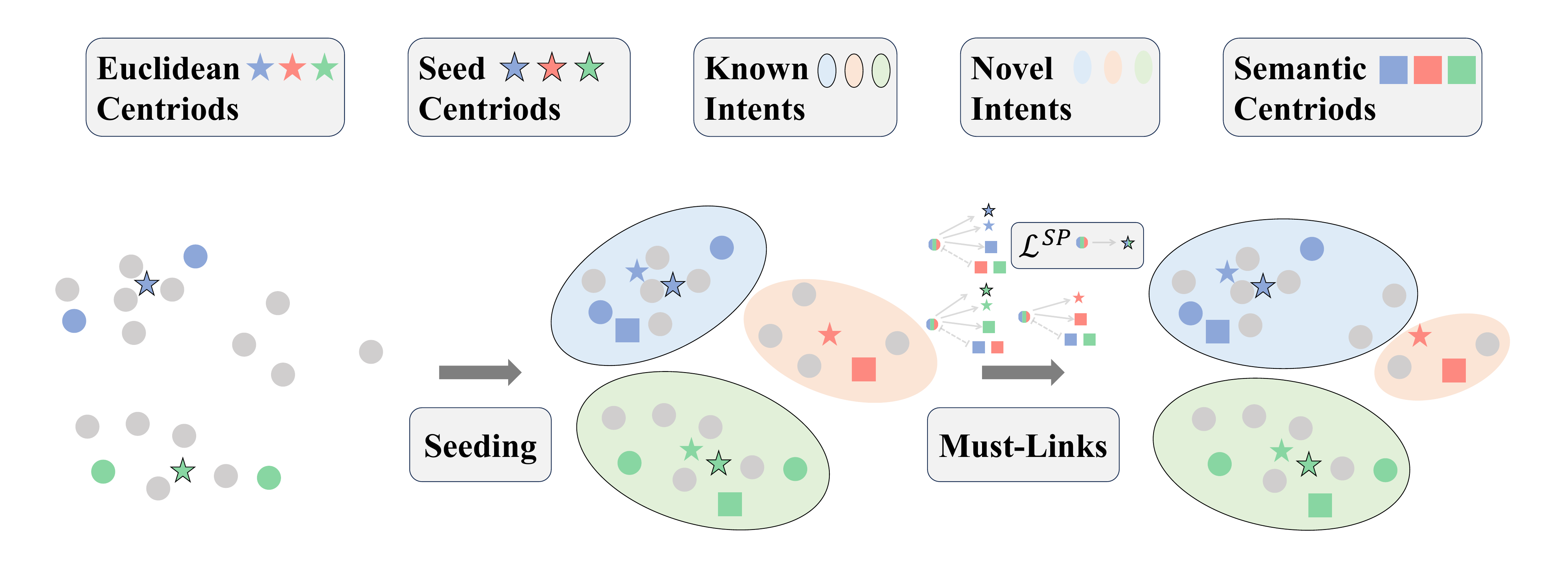} 
    \vspace{-6ex}
    \caption{Semi-Supervised Optimizations of \algo.}
    \label{fig:nid-semi-injection}
\vspace{-2ex}
\end{figure}

Recall that under semi-supervised settings, we are provided with a limited set $\D_{l}$ of labeled utterance samples. In previous works, such supervision signals are solely considered in the text encoding phase, which are often underexploited.
In semi-supervised NID, in addition to applying the PTE finetuned on $\mathcal{D}_{\text{train}}$ containing both labeled and unlabeled datasets, \algo{} injects ancillary knowledge from $\D_{l}$ to the clustering stage through the {\em seeding} and {\em soft must-links} (SML) techniques in the sequel. The high-level ideas of these two optimizations are illustrated in Fig. \ref{fig:nid-semi-injection}.

\stitle{Seeding}
Taking inspiration from {\em seeded $K$-Means}~\cite{basu2002semi}, we perform a warm-start for the clustering process by aligning a subset of the initial Euclidean centroids with the known intents $\Y_k$ from the labeled dataset $\mathcal{D}_{l}$. The initial Euclidean centroids, denoted as $\{\boldsymbol{\mu}_k^{0}\}_{k=1}^K$, are obtained by initially running $K$-\texttt{Means++} over all input text embeddings (Line 2 in Algorithm~\ref{alg:nilc}). 
The seed centroids, $\{\boldsymbol{\mu}^*_j\}_{j=1}^M$, are derived from the labeled data $\mathcal{D}_l$ by computing the mean embedding for all utterances belonging to each of the $M$ known intents in $\Y_k$, i.e.,
\begin{equation*}
\boldsymbol{\mu}^*_j = \textsf{Mean}(\{\xvec_i|(x_i,y_i)\in \mathcal{D}_l, y_i=j\}).
\end{equation*}

The matching from $\{\boldsymbol{\mu}_k^{0}\}_{k=1}^K$ to $\{\boldsymbol{\mu}^*_j\}_{j=1}^M$ is framed as a linear assignment problem. The objective is to find an optimal mapping $\pi$ that minimizes the total semantic distance, i.e., cosine dissimilarity, between the seed centroids and the initial Euclidean centroids:
\begin{equation}
\min_{\pi} \sum_{j=1}^{M} \left(1 - \text{cos}(\boldsymbol{\mu}^*_j, \boldsymbol{\mu}_{\pi(j)}^{0})\right),
\end{equation}
which can be readily solved using the {\em Hungarian algorithm}~\cite{mills2007dynamic}.

Once the mapping $\pi$ is found, the $M$ initial centroids are replaced by their corresponding seed centroids: $\boldsymbol{\mu}_{\pi(j)}^{0} \leftarrow \boldsymbol{\mu}^*_j$. This procedure anchors a portion of the clusters in the known semantic space, providing a strong inductive bias from the outset.

\stitle{Soft Must-Links}
Inspired by the principles of {\em constrained clustering}~\cite{basu2004active,wang2014constrained}, we propose to impose constraints in the form of soft must-links in the course of iterative updating clusters. Such soft must-links aim at pulling utterance samples towards clusters that have been mapped to their known intents.
Unlike previous constrained clustering methods~\cite{basu2004active,wang2014constrained} that rely on pre-defined and static pairwise constraints, our approach is dynamic and semantically driven, providing a weighted and soft pull towards known intents.

This process involves two main steps. Firstly, in the $t$-th iteration, a dynamic one-to-one mapping $\pi^{t}$ is established between the $M$ known intents and the current $K$ clusters. 
The mapping can be built based on either the similarities between their centroid embeddings or the answers from LLMs. Due to space limit, we defer the details and evaluations for these two strategies to Appendix~\ref{sec:map-strategy}, 
\ref{sec:case-study-mapping}, and 
\ref{sec:empiracal-study-mapping-sampling}. 
After that, if a cluster $\C_k$ is mapped to a known intent $j$, i.e., $\pi^{t}(k) = j$, we introduce an extra term to our cost function $f(\xvec_i)$ in Eq.~\eqref{eq:clust-cost-i} for assigning clusters as supervision:
\begin{equation}
\mathcal{L}^{\text{SP}}_i = 1 - \text{cos}(\xvec_i, \boldsymbol{\mu}^*_{\pi^{t}(k)}),
\end{equation}
which seeks to minimize the distance from sample $\xvec_i$ to the known intent cluster $\boldsymbol{\mu}^*_{\pi^{t}(k)}$, acting as a soft must-link between them. Consequently, this leads to our new cost function:
\begin{equation}
f'(\xvec_i) = \mathcal{L}^{\text{ED}}_i + \alpha \cdot \mathcal{L}^{\text{SC}}_i + \beta \cdot \mathcal{L}^{\text{SS}}_i + \gamma \cdot \mathcal{L}^{\text{SP}}_i,
\end{equation}
where the hyperparameter $\gamma$ controls the strength of these soft must-link. 
Note that if no mapping exists for cluster $\C_k$, the cost function remains the same in Eq.~\eqref{eq:clust-cost-i}.

\subsection{Analysis}\label{sec:analysis}
The iterative process of our framework is analogous to an Expectation-Maximization (EM) algorithm, seeking to find an optimal partition $\C = \{\C_k\}_{k=1}^K$ and its associated representations (centroids $\{\boldsymbol{\mu}_k, \boldsymbol{\theta}_k\}$ and potentially refined embeddings $\{\tilde{\xvec}_i\}$) that minimize the global objective function $\mathcal{L} = \sum_{k=1}^{K} \sum_{\xvec_i \in \C_k} f'(\xvec_i)$. Our algorithm iteratively seeks a local minimum for this objective by alternating between assignment and update/refinement phases.

The $T$ iterations of \algo{} are performed periodically at specified intervals, while most other iterations are standard $K$-\texttt{Means} steps. Regarding computational cost, most standard steps consisting of only $K$-\texttt{Means} have a complexity of $\mathcal{O}(NKd)$ dominated by the assignment phase, on which a complete \algo{} iteration builds with the same primary complexity of $\mathcal{O}(NKd)$, adding overhead from a fixed number of LLM calls ($K+\delta$) and the optional Hungarian algorithm for mapping ($\mathcal{O}(\max(M, K)^3)$). Since $K, M, \delta \ll N$, the $\mathcal{O}(NKd)$ term remains the dominant factor, and the computational cost of \algo{} is competitive with standard $K$-\texttt{Means}.

\section{Experiments}
This section experimentally evaluates the effectiveness of the proposed \algo{} framework. All experiments are conducted on a Linux machine with an NVIDIA A100 GPU (80GB RAM), AMD EPYC 7513 CPU (2.6 GHz), and 1TB RAM. The source code and datasets are publicly accessible at \url{https://github.com/HKBU-LAGAS/NILC}.

\subsection{Datasets}
\begin{table}[!t]
\centering
\caption{Dataset statistics.}
\vspace{-2ex}
\renewcommand{\arraystretch}{1.0}
\resizebox{\columnwidth}{!}{
\begin{small}
\begin{tabular}{c|c|c|c}
\hline
{\bf Dataset} & {\bf \#Utterances} ($N$) & {\bf \#Intents} ($K$) & {\bf Domain} \\ \hline
CLINC & 22,500 & 150  & General \\ 
BANKING  & 13,083  & 77  &  Banking\\ 
StackOverflow  & 20,000  & 20  &  Technical \\ 
M-CID & 1,745  & 16  & Covid-19 \\ 
SNIPS & 14,484  & 7  &  Voice \\
DBPedia & 14,000  & 14  &  Ontology \\
\hline
\end{tabular}
\end{small}
}
\label{tab:datasets}
\vspace{-2ex}
\end{table}

We conduct extensive experiments on six challenging benchmark datasets to evaluate the performance of our proposed method. The datasets include:
CLINC, a multi-domain dataset with 22,500 utterances and 150 intents;
BANKING, a fine-grained dataset from the banking domain with 13,083 utterances and 77 intents;
StackOverflow, a technical question dataset with 20,000 samples across 20 classes;
SNIPS, a personal voice assistant dataset containing 14,484 utterances with 7 intents;
M-CID, which consists of 1,745 utterances related to 16 COVID-19 service intents;
and DBPedia, a dataset of 14,000 samples from 14 non-overlapping ontology classes.
A summary of the dataset statistics is provided in Table~\ref{tab:datasets}.

\subsection{Baselines and Settings}
We compare our method against a wide range of baselines.

\noindent\textbf{Unsupervised Methods.} This category does not use any labeled data. We compare against: \SAE~\cite{vincent2010stacked}, which is based on a stacked autoencoder; \DEC~\cite{xie2016unsupervised} and \DCN~\cite{shen2021semi}, which are deep clustering frameworks; \CC~\cite{li2021contrastive} and \SCCL~\cite{zhang2021supporting}, which are based on contrastive learning; and \USNID~\cite{zhang2023clustering}, which leverages pre-training for intent discovery.

\noindent\textbf{Semi-Supervised Methods.} These methods leverage a small amount of labeled data. This category includes methods from various technical approaches: constrained clustering (\KCL~\cite{hsu2017learning}, \MCL~\cite{hsu2019multi}); novel category discovery adapted from computer vision (\DTC~\cite{han2019learning}, \GCD~\cite{vaze2022generalized}); and methods for new intent discovery (\CDAC~\cite{lin2020discovering}, \DeepAligned~\cite{zhang2021discovering}, \SDC~\cite{an2025unleashing}, \MTPCLNN~\cite{zhang2022new}, \LatentEM~\cite{zhou2023probabilistic}). We also include \LANID~\cite{fan2025lanid}, which uses an LLM to generate pairwise relationship labels for contrastive fine-tuning; and \IntentGPT~\cite{rodriguez2024intentgpt}, which employs the LLM as a few-shot discoverer through a sophisticated prompting strategy. Additionally, we evaluate \USNID in its semi-supervised setting.

Unless specified otherwise, we employ \USNID as $\textsf{PTE}(\cdot)$ and \GPTMini as $\textsf{LLM}(\cdot)$ in \algo{}.
The detailed settings for hyperparameters are provided in Appendix~\ref{sec:param-set}.
Following prior studies~\cite{zhang2022new,zhang2023clustering,fan2025lanid,rodriguez2024intentgpt}, {\em normalized mutual information} (NMI), {\em adjusted rand score} (ARI), and {\em clustering accuracy} (ACC) are used as NID metrics.

\subsection{NID Performance}

Table~\ref{tab:performance_main} reports the NMI, ARI, and ACC scores for \algo{} and all baseline methods on the six benchmark datasets. We can draw several key observations.

First, under the unsupervised setting, \algo{} consistently and significantly outperforms all baselines across all datasets. For instance, on the M-CID dataset, \algo{} achieves an improvement of 7.22\% in NMI, 7.72\% in ARI, and 11.38\% in ACC over the strongest baseline, \USNID. This demonstrates the effectiveness of our LLM-assisted DCS and HSR in discovering coherent intent clusters without any labeled data.

Second, in the more practical semi-supervised setting, \algo{} continues to establish its superiority. It surpasses all recent and competitive baselines, including those that also leverage LLMs like \LANID and \IntentGPT. On the CLINC dataset, for example, \algo{} improves upon the runner-up \USNID by 0.41\% in NMI, 1.76\% in ARI, and 1.67\% in ACC. These gains are consistent across diverse domains, from general intents in CLINC to technical questions in StackOverflow, validating the robustness of injecting supervised signals through IS and ML within our iterative framework.

Another important observation is that while methods based on deep representation learning (e.g., \USNID, \MTPCLNN) form a strong baseline, our framework's ability to iteratively refine both cluster assignments and embeddings provides an additional performance boost. This highlights the limitations of a static, cascaded approach and confirms the benefits of a more synergistic methodology where the clustering process and embedding space mutually enhance each other with the aid of an LLM's reasoning capabilities.

\begin{table}[!t]
\centering
\renewcommand{\arraystretch}{1.1} 
\caption{Ablation study on \algo.}
\label{tab:ablation}
\vspace{-2ex}
\resizebox{\columnwidth}{!}{%
\begin{tabular}{l c c c}
\toprule
\multirow{2}{*}{Variant} 
& DBPedia & M-CID & StackOverflow \\
\cmidrule(lr){2-4} 
& NMI/ARI/ACC & NMI/ARI/ACC & NMI/ARI/ACC \\
\midrule
\algo (unsup.)
& 78.21/65.61/73.43 & 71.87/52.36/68.77 & 72.08/64.35/76.62 \\
w/o DCS
& 77.14/64.25/72.00 & 69.49/49.25/67.34 &  71.67/63.95/76.45 \\
w/o HSR
& 77.49/64.87/72.71 & 69.89/49.93/67.62 & 71.80/63.93/76.45 \\
\midrule
\algo (semi.)
& 89.99/84.88/92.00 & 83.42/73.20/85.10 & 80.73/76.67/87.28 \\
w/o DCS
& 88.61/83.12/91.00 & 80.75/69.33/81.95 & 80.56/75.42/86.60 \\
w/o HSR
& 88.73/83.43/91.14 & 81.42/70.37/82.81 & 80.57/76.09/86.98 \\
w/o Seeding
& 89.36/84.19/91.57 & 82.20/71.73/83.95 & 80.53/76.36/87.08 \\
w/o SML
& 89.09/83.87/91.43 & 82.06/71.33/83.67 &  80.28/76.16/86.95 \\
\bottomrule
\end{tabular}
}
\vspace{0ex}
\end{table}

\begin{table*}[!t]
\centering
\setlength{\tabcolsep}{3pt} 
\renewcommand{\arraystretch}{1.10}
\caption{NID Performance comparison. (best \textbf{bolded} and runner-up \underline{underlined})}
\label{tab:performance_main}
\vspace{-2ex}
\begin{small}
\resizebox{\linewidth}{!}{
\begin{tabular}{l l c c c c c c}
    \toprule
    & \multirow{2}{*}{Method} & CLINC & BANKING & StackOverflow & M-CID & SNIPS & DBPedia \\
    & & NMI/ARI/ACC & NMI/ARI/ACC & NMI/ARI/ACC & NMI/ARI/ACC & NMI/ARI/ACC & NMI/ARI/ACC \\
    \midrule
    \multirow{8}{*}{\rotatebox[origin=c]{90}{Unsupervised}} 
    & \SAE & 74.07/32.06/47.80 & 60.01/24.17/37.94 & 46.35/29.65/51.62 & 50.49/43.61/53.07 & 76.07/69.80/81.63 & 71.34/57.57/70.07 \\
    & \DEC & 75.14/32.22/49.24 & 62.85/25.94/38.84 & 60.26/36.92/59.93 & 51.09/\underline{44.64}/53.73 & 84.49/80.58/87.80 & 74.54/59.87/70.74 \\
    & \DCN & 75.15/32.20/49.23 & 62.81/25.92/38.83 & 60.41/37.02/60.00 & 51.09/\underline{44.64}/53.73 & 85.52/80.61/87.80 & 74.57/\underline{59.89}/\underline{70.76} \\
    & \CC & 66.05/18.34/33.09 & 44.64/9.73/21.21 & 20.38/9.21/21.99 & 55.75/33.08/50.29 & 82.96/77.02/85.90 & 71.56/53.29/66.79 \\
    & \SCCL & 79.14/38.12/49.96 & 63.43/26.32/39.92 & 68.69/36.97/68.28 & 55.18/30.05/48.71 & 72.88/55.53/68.81 & \underline{77.02}/59.29/67.57 \\
    & \USNID & \underline{91.45}/\underline{70.02}/\underline{77.06} & \underline{75.61}/\underline{43.96}/\underline{55.09} & \underline{71.49}/\underline{52.13}/\underline{69.20} & \underline{64.65}/41.51/\underline{57.39} & \underline{89.46}/\underline{85.72}/\underline{91.43} & 75.15/58.90/68.06 \\ \cline{2-8}
    & \algo{} & \textbf{91.58}/\textbf{71.53}/\textbf{78.36} & \textbf{77.43}/\textbf{47.74}/\textbf{59.45} & \textbf{72.08}/\textbf{64.35}/\textbf{76.62} & \textbf{71.87}/\textbf{52.36}/\textbf{68.77} & \textbf{91.56}/\textbf{90.34}/\textbf{95.57} & \textbf{78.21}/\textbf{65.61}/\textbf{73.43} \\ 
    & Improv. & +0.13/+1.51/+1.30 & +1.82/+3.78/+4.36 & +0.59/+12.22/+7.42 & +7.22/+7.72/+11.38 & +2.10/+4.62/+4.14 & +1.19/+5.72/+2.67 \\ 
    \midrule
    \midrule
    \multirow{14}{*}{\rotatebox[origin=c]{90}{Semi-Supervised}} 
    & \KCL & 86.10/58.86/69.22 & 73.07/45.49/59.27 & 64.84/55.86/71.06 & 44.10/22.29/40.92 & 77.94/64.11/74.00 & 78.78/61.63/69.13 \\
    & \MCL & 87.38/61.48/70.39 & 74.46/48.07/61.52 & 63.24/55.97/71.31 & 55.25/35.06/53.44 & 81.07/67.75/74.44 & 79.94/64.22/72.73 \\
    & \DTC & 89.43/67.26/77.61 & 73.98/43.95/56.11 & 62.64/53.32/70.36 & 33.78/11.43/29.80 & 73.50/62.83/74.01 & 77.72/58.98/68.14 \\
    & \CDAC & 85.93/55.81/68.01 & 68.03/35.61/48.77 & 55.85/41.82/62.53 & 55.64/32.97/52.92 & 83.13/77.36/86.97 & 80.23/65.38/75.34 \\
    & \GCD & 88.99/65.58/76.42 & 71.99/42.85/56.43 & 60.80/42.25/65.28 & 60.71/40.80/58.71 & 81.52/78.10/89.72 & 79.36/64.81/76.63 \\
    & \DeepAligned & 93.89/79.75/86.49 & 79.12/52.46/63.73 & 73.83/60.26/77.87 & 48.34/23.28/41.26 & 88.09/85.21/92.71 & 84.34/69.99/78.86 \\
    & \MTPCLNN & 95.44/84.23/89.30 & 84.63/64.32/75.33 & 73.88/64.04/79.68 & 76.76/63.29/78.08 & 89.95/87.90/93.39 & 80.17/67.13/78.14 \\
    & \LatentEM & 94.86/82.40/88.40 & 81.81/57.96/70.42 & 75.46/63.30/74.30 & 68.40/48.12/63.61 & 82.89/76.32/83.43 & \textbf{91.46}/\underline{83.40}/\underline{87.00} \\
    & \USNID & \underline{96.46}/\underline{86.49}/\underline{90.20} & \underline{87.67}/\underline{69.93}/\underline{78.52} & \underline{80.01}/\underline{74.74}/\underline{85.61} & 79.04/66.53/78.83 & \underline{93.32}/\underline{91.94}/\underline{96.28} & 86.29/76.79/85.40 \\
    & \SDC & 95.33/83.85/89.32 & 85.11/66.16/77.22 & 77.45/60.10/80.25 & 46.69/22.13/41.52 & 89.04/86.57/94.04 & 80.03/67.41/81.53 \\
    & \IntentGPT$^{\diamondsuit}$ & 96.06/84.76/88.76 & 85.94/66.66/77.21 & - & - & - & - \\ 
    & \LANID & 96.08/85.25/89.81 & 87.18/68.56/76.75 & 75.30/64.71/77.42 & \underline{82.53}/\underline{69.84}/\underline{81.09} & 91.70/90.23/94.75 & 85.38/74.72/83.87 \\ \cline{2-8}
    & \algo{} & \textbf{96.87}/\textbf{88.25}/\textbf{91.87} & \textbf{87.74}/\textbf{71.30}/\textbf{81.07} & \textbf{80.73}/\textbf{76.67}/\textbf{87.28} & \textbf{83.42}/\textbf{73.20}/\textbf{85.10} & \textbf{95.61}/\textbf{95.14}/\textbf{97.86} & \underline{89.99}/\textbf{84.88}/\textbf{92.00} \\ 
    & Improv. & +0.41/+1.76/+1.67 & +0.07/+1.37/+2.55 & +0.72/+1.93/+1.67 & +0.89/+3.36/+4.01 & +2.29/+3.20/+1.58 & -1.47/+1.48/+5.00 \\
    \bottomrule
\end{tabular}
}
\vspace{0ex}
$^{\diamondsuit}$Results are taken from the original paper. Missing values (-) indicate that the results are not available.
\end{small}
\end{table*}

\begin{table}[!t]
\centering
\caption{Performance comparison of \algo on various encoders across different datasets.}
\label{tab:performance_encoders}
\vspace{-2ex}
\begin{small}
\resizebox{\columnwidth}{!}{
\begin{tabular}{l c c c}
    \toprule
    \multirow{2}{*}{Method} & DBPedia & M-CID & StackOverflow \\
    & (NMI/ARI/ACC) & (NMI/ARI/ACC) & (NMI/ARI/ACC) \\
    \midrule
    \SentenceBERT         &  70.34/54.58/67.71 & 60.30/37.19/59.03 & 68.08/55.67/67.32 \\
    {\algo{} (\SentenceBERT)} & 71.29/56.85/72.86 & 67.89/39.17/65.04 & 70.82/57.35/69.08 \\
    \midrule
    \Instructor      & 70.79/57.65/71.86 & 56.46/31.57/49.28 & 74.21/62.81/77.28 \\
    {\algo{} (\Instructor)} & 77.70/64.66/73.43 & 62.55/40.62/59.03 &  75.78/63.00/77.40 \\
    \midrule
    \MTPCLNN     & 80.17/67.13/78.14 & 76.76/63.29/78.08 & 73.88/64.04/79.68 \\
    {\algo{} (\MTPCLNN)} & 81.23/68.42/79.56 & 77.18/64.23/78.85 & 74.69/65.27/80.54 \\
    \midrule
    \LatentEM     & 91.46/83.40/87.00 & 68.40/48.12/63.61 & 75.46/63.30/74.30 \\
    {\algo{} (\LatentEM)} & 91.52/85.61/89.75 & 69.14/51.24/64.97 & 76.34/64.29/75.72 \\
    \bottomrule
\end{tabular}
}
\end{small}
\end{table}

\begin{figure}[!t]
\centering
\begin{small}
\subfloat[DBPEDIA]{
    \begin{tikzpicture}[scale=1]
    \begin{axis}[
        height=\columnwidth/2.7,
        width=\columnwidth/2.5,
        ylabel={ANA},
        xmin=0.0, xmax=1.0,
        ymin=87.4, ymax=88.6,
        xtick={0.1, 0.3, 0.5, 0.7, 0.9},
        ytick={87.50, 87.75, 88.00, 88.25, 88.50}, %
        yticklabel style={
            font=\scriptsize,
            /pgf/number format/fixed,
            /pgf/number format/precision=2,
            /pgf/number format/fixed zerofill
        },
        yticklabel style = {font=\tiny},
        every axis label/.style={font=\footnotesize},
        title style={font=\footnotesize},
        grid=major,
        grid style={dashed, gray!30}
    ]
    \addplot[line width=0.4mm, mark=o, color=blue!80!black]
        plot coordinates {
            (0.1, 88.01) (0.3, 88.30) (0.5, 87.86) (0.7, 87.58) (0.9, 88.16)
        };
    \end{axis}
    \end{tikzpicture}
}
\subfloat[MCID]{
    \begin{tikzpicture}[scale=1]
    \begin{axis}[
        height=\columnwidth/2.7,
        width=\columnwidth/2.5,
        xmin=0.0, xmax=1.0,
        ymin=76.7, ymax=80.3,
        xtick={0.1, 0.3, 0.5, 0.7, 0.9},
        ytick={77.00, 77.75, 78.50, 79.25, 80.00},
        yticklabel style={
            font=\scriptsize,
            /pgf/number format/fixed,
            /pgf/number format/precision=2,
            /pgf/number format/fixed zerofill
        },
        yticklabel style = {font=\tiny},
        every axis label/.style={font=\footnotesize},
        title style={font=\footnotesize},
        grid=major,
        grid style={dashed, gray!30}
    ]
    \addplot[line width=0.4mm, mark=o, color=blue!80!black]
        plot coordinates {
            (0.1, 79.55) (0.3, 79.32) (0.5, 79.29) (0.7, 78.66) (0.9, 77.34)
        };
    \end{axis}
    \end{tikzpicture}
}
\subfloat[StackOverflow]{
    \begin{tikzpicture}[scale=1]
    \begin{axis}[
        height=\columnwidth/2.7,
        width=\columnwidth/2.5,
        xmin=0.0, xmax=1.0,
        ymin=80.74, ymax=81.46,
        xtick={0.1, 0.3, 0.5, 0.7, 0.9},
        ytick={80.80, 80.95, 81.10, 81.25, 81.40},
        yticklabel style={
            font=\scriptsize,
            /pgf/number format/fixed,
            /pgf/number format/precision=2,
            /pgf/number format/fixed zerofill
        },
        yticklabel style = {font=\tiny},
        every axis label/.style={font=\footnotesize},
        title style={font=\footnotesize},
        grid=major,
        grid style={dashed, gray!30}
    ]
    \addplot[line width=0.4mm, mark=o, color=blue!80!black]
        plot coordinates {
            (0.1, 80.86) (0.3, 80.92) (0.5, 81.21) (0.7, 81.03) (0.9, 81.24)
        };
    \end{axis}
    \end{tikzpicture}
}
\end{small}
\vspace{-2ex}
\caption{Varying \(\alpha\).}
\label{fig:hyperparameter_alpha}
\vspace{-2ex}
\end{figure}
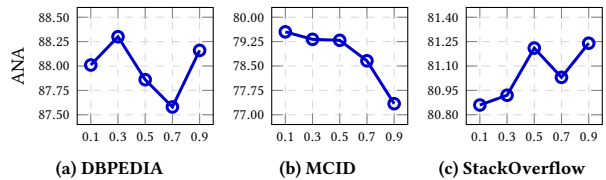

\begin{figure}[!t]
\centering
\begin{small}
\subfloat[DBPEDIA]{
    \begin{tikzpicture}[scale=1]
    \begin{axis}[
        height=\columnwidth/2.7,
        width=\columnwidth/2.5,
        ylabel={ANA},
        xmin=0.0, xmax=1.0,
        ymin=87.4, ymax=88.6,
        xtick={0.1, 0.3, 0.5, 0.7, 0.9},
        ytick={87.50, 87.75, 88.00, 88.25, 88.50},
        yticklabel style={
            font=\scriptsize,
            /pgf/number format/fixed,
            /pgf/number format/precision=2,
            /pgf/number format/fixed zerofill
        },
        yticklabel style = {font=\tiny},
        every axis label/.style={font=\footnotesize},
        title style={font=\footnotesize},
        grid=major,
        grid style={dashed, gray!30}
    ]
    \addplot[line width=0.4mm, mark=square, color=green!60!black]
        plot coordinates {
            (0.1, 88.10) (0.3, 87.86) (0.5, 87.86) (0.7, 88.07) (0.9, 88.07)
        };
    \end{axis}
    \end{tikzpicture}
}
\subfloat[MCID]{
    \begin{tikzpicture}[scale=1]
    \begin{axis}[
        height=\columnwidth/2.7,
        width=\columnwidth/2.5,
        xmin=0.0, xmax=1.0,
        ymin=78.3, ymax=80.7,
        xtick={0.1, 0.3, 0.5, 0.7, 0.9},
        ytick={78.50, 79.00, 79.50, 80.00, 80.50},
        yticklabel style={
            font=\scriptsize,
            /pgf/number format/fixed,
            /pgf/number format/precision=2,
            /pgf/number format/fixed zerofill
        },
        yticklabel style = {font=\tiny},
        every axis label/.style={font=\footnotesize},
        title style={font=\footnotesize},
        grid=major,
        grid style={dashed, gray!30}
    ]
    \addplot[line width=0.4mm, mark=square, color=green!60!black]
        plot coordinates {
            (0.1, 79.93) (0.3, 79.67) (0.5, 79.29) (0.7, 79.33) (0.9, 78.79)
        };
    \end{axis}
    \end{tikzpicture}
}
\subfloat[StackOverflow]{
    \begin{tikzpicture}[scale=1]
    \begin{axis}[
        height=\columnwidth/2.7,
        width=\columnwidth/2.5,
        xmin=0.0, xmax=1.0,
        ymin=80.52, ymax=81.48,
        xtick={0.1, 0.3, 0.5, 0.7, 0.9},
        ytick={80.60, 80.80, 81.00, 81.20, 81.40},
        yticklabel style={
            font=\scriptsize,
            /pgf/number format/fixed,
            /pgf/number format/precision=2,
            /pgf/number format/fixed zerofill
        },
        yticklabel style = {font=\tiny},
        every axis label/.style={font=\footnotesize},
        title style={font=\footnotesize},
        grid=major,
        grid style={dashed, gray!30}
    ]
    \addplot[line width=0.4mm, mark=square, color=green!60!black]
        plot coordinates {
            (0.1, 80.95) (0.3, 80.76) (0.5, 81.21) (0.7, 80.96) (0.9, 81.09)
        };
    \end{axis}
    \end{tikzpicture}
}
\end{small}
\vspace{-2ex}
\caption{Varying $\beta$.}
\label{fig:hyperparameter_beta}
\vspace{-2ex}
\end{figure}
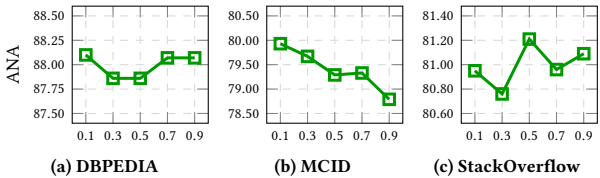

\begin{figure}[!t]
\centering
\begin{small}
\subfloat[DBPEDIA]{
    \begin{tikzpicture}[scale=1]
    \begin{axis}[
        height=\columnwidth/2.7,
        width=\columnwidth/2.5,
        ylabel={ANA},
        xmin=0.0, xmax=1.0,
        ymin=87.4, ymax=88.6,
        xtick={0.1, 0.3, 0.5, 0.7, 0.9},
        ytick={87.50, 87.75, 88.00, 88.25, 88.50},
        yticklabel style={
            font=\scriptsize,
            /pgf/number format/fixed,
            /pgf/number format/precision=2,
            /pgf/number format/fixed zerofill
        },
        yticklabel style = {font=\tiny},
        every axis label/.style={font=\footnotesize},
        title style={font=\footnotesize},
        grid=major,
        grid style={dashed, gray!30}
    ]
    \addplot[line width=0.4mm, mark=triangle, color=red!80!black]
        plot coordinates {
            (0.1, 88.13) (0.3, 87.86) (0.5, 87.86) (0.7, 87.77) (0.9, 88.37)
        };
    \end{axis}
    \end{tikzpicture}
}
\subfloat[MCID]{
    \begin{tikzpicture}[scale=1]
    \begin{axis}[
        height=\columnwidth/2.7,
        width=\columnwidth/2.5,
        xmin=0.0, xmax=1.0,
        ymin=77.8, ymax=80.2,
        xtick={0.1, 0.3, 0.5, 0.7, 0.9},
        ytick={78.00, 78.50, 79.00, 79.50, 80.00},
        yticklabel style={
            font=\scriptsize,
            /pgf/number format/fixed,
            /pgf/number format/precision=2,
            /pgf/number format/fixed zerofill
        },
        yticklabel style = {font=\tiny},
        every axis label/.style={font=\footnotesize},
        title style={font=\footnotesize},
        grid=major,
        grid style={dashed, gray!30}
    ]
    \addplot[line width=0.4mm, mark=triangle, color=red!80!black]
        plot coordinates {
            (0.1, 78.27) (0.3, 78.26) (0.5, 79.29) (0.7, 80.02) (0.9, 78.73)
        };
    \end{axis}
    \end{tikzpicture}
}
\subfloat[StackOverflow]{
    \begin{tikzpicture}[scale=1]
    \begin{axis}[
        height=\columnwidth/2.7,
        width=\columnwidth/2.5,
        xmin=0.0, xmax=1.0,
        ymin=80.72, ymax=81.68,
        xtick={0.1, 0.3, 0.5, 0.7, 0.9},
        yticklabel style={
            font=\scriptsize,
            /pgf/number format/fixed,
            /pgf/number format/precision=2,
            /pgf/number format/fixed zerofill
        },
        yticklabel style = {font=\tiny},
        every axis label/.style={font=\footnotesize},
        title style={font=\footnotesize},
        grid=major,
        grid style={dashed, gray!30}
    ]
    \addplot[line width=0.4mm, mark=triangle, color=red!80!black]
        plot coordinates {
            (0.1, 81.39) (0.3, 81.28) (0.5, 81.21) (0.7, 80.90) (0.9, 80.91)
        };
    \end{axis}
    \end{tikzpicture}
}
\end{small}
\vspace{-2ex}
\caption{Varying \(\gamma\).}
\label{fig:hyperparameter_gamma}
\vspace{-2ex}
\end{figure}
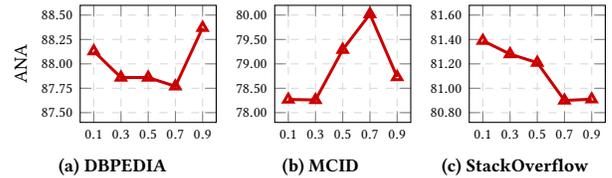

\begin{figure}[!t]
\centering
\begin{small}
\subfloat[DBPEDIA]{
	\begin{tikzpicture}[scale=1]
	\begin{axis}[
		height=\columnwidth/2.7,
		width=\columnwidth/2.5,
		ylabel={ANA},
		xmin=0, xmax=30,
		ymin=87.92, ymax=88.88,
		xtick={5,10,15,20,25},
		ytick={88.00, 88.20, 88.40, 88.60, 88.80},
		yticklabel style={
			font=\scriptsize,
			/pgf/number format/fixed,
			/pgf/number format/precision=2,
			/pgf/number format/fixed zerofill
		},
		yticklabel style = {font=\tiny},
		every axis label/.style={font=\tiny},
		title style={font=\footnotesize},
		grid=major,
		grid style={dashed, gray!30}
	]
	\addplot[line width=0.4mm, mark=diamond, color=orange!80!black]
		plot coordinates {
			(5, 88.62) (10, 88.37) (15, 88.13) (20, 88.40) (25, 88.53)
		};
	\end{axis}
	\end{tikzpicture}
}
\subfloat[MCID]{
	\begin{tikzpicture}[scale=1]
	\begin{axis}[
		height=\columnwidth/2.7,
		width=\columnwidth/2.5,
		xmin=0, xmax=30,
		ymin=77.45, ymax=81.05,
		xtick={5,10,15,20,25},
		ytick={77.75, 78.50, 79.25, 80.00, 80.75},
		yticklabel style={
			font=\scriptsize,
			/pgf/number format/fixed,
			/pgf/number format/precision=2,
			/pgf/number format/fixed zerofill
		},
		yticklabel style = {font=\tiny},
		every axis label/.style={font=\tiny},
		title style={font=\footnotesize},
		grid=major,
		grid style={dashed, gray!30}
	]
	\addplot[line width=0.4mm, mark=diamond, color=orange!80!black]
		plot coordinates {
			(5, 78.01) (10, 80.02) (15, 80.57) (20, 79.36) (25, 78.88)
		};
	\end{axis}
	\end{tikzpicture}
}
\subfloat[StackOverflow]{
	\begin{tikzpicture}[scale=1]
	\begin{axis}[
		height=\columnwidth/2.7,
		width=\columnwidth/2.5,
		xmin=0, xmax=30,
		ymin=81.288, ymax=81.432,
		xtick={5,10,15,20,25},
		ytick={81.30, 81.33, 81.36, 81.39, 81.42},
		yticklabel style={
			font=\scriptsize,
			/pgf/number format/fixed,
			/pgf/number format/precision=2,
			/pgf/number format/fixed zerofill
		},
		yticklabel style = {font=\tiny},
		every axis label/.style={font=\tiny},
		title style={font=\footnotesize},
		grid=major,
		grid style={dashed, gray!30}
	]
	\addplot[line width=0.4mm, mark=diamond, color=orange!80!black]
		plot coordinates {
			(5, 81.35) (10, 81.39) (15, 81.32) (20, 81.37) (25, 81.32)
		};
	\end{axis}
	\end{tikzpicture}
}
\end{small}
\vspace{-2ex}
\caption{Varying the number \(\delta\) of hard samples.}
\label{fig:hyperparameter_analysis_unique_style_O}
\vspace{-2ex}
\end{figure}
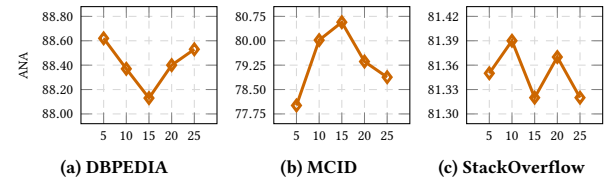

\begin{figure}[!t]
\centering
\begin{small}
\subfloat[DBPEDIA]{
	\begin{tikzpicture}[scale=1]
	\begin{axis}[
		height=\columnwidth/2.7,
		width=\columnwidth/2.5,
		ylabel={ANA},
		xmin=0, xmax=12,
		ymin=88.068, ymax=88.452,
		xtick={2, 4, 6, 8, 10},
		ytick={88.10, 88.18, 88.26, 88.34, 88.42},
		yticklabel style={
			font=\scriptsize,
			/pgf/number format/fixed,
			/pgf/number format/precision=2,
			/pgf/number format/fixed zerofill
		},
		yticklabel style = {font=\tiny},
		every axis label/.style={font=\tiny},
		title style={font=\footnotesize},
		grid=major,
		grid style={dashed, gray!30}
	]
	\addplot[line width=0.4mm, mark=*, color=violet!80!black]
		plot coordinates {
			(2, 88.15) (4, 88.38) (6, 88.18) (8, 88.37) (10, 88.37)
		};
	\end{axis}
	\end{tikzpicture}
}
\subfloat[MCID]{
	\begin{tikzpicture}[scale=1]
	\begin{axis}[
		height=\columnwidth/2.7,
		width=\columnwidth/2.5,
		xmin=0, xmax=12,
		ymin=77.24, ymax=80.36,
		xtick={2, 4, 6, 8, 10},
		ytick={77.50, 78.15, 78.80, 79.45, 80.10},
		yticklabel style={
			font=\scriptsize,
			/pgf/number format/fixed,
			/pgf/number format/precision=2,
			/pgf/number format/fixed zerofill
		},
		yticklabel style = {font=\tiny},
		every axis label/.style={font=\tiny},
		title style={font=\footnotesize},
		grid=major,
		grid style={dashed, gray!30}
	]
	\addplot[line width=0.4mm, mark=*, color=violet!80!black]
		plot coordinates {
			(2, 78.62) (4, 77.71) (6, 78.20) (8, 78.80) (10, 80.02)
		};
	\end{axis}
	\end{tikzpicture}
}
\subfloat[StackOverflow]{
	\begin{tikzpicture}[scale=1]
	\begin{axis}[
		height=\columnwidth/2.7,
		width=\columnwidth/2.5,
		xmin=0, xmax=12,
		ymin=81.18, ymax=81.42,
		xtick={2, 4, 6, 8, 10},
		ytick={81.20, 81.25, 81.30, 81.35, 81.40},
		yticklabel style={
			font=\scriptsize,
			/pgf/number format/fixed,
			/pgf/number format/precision=2,
			/pgf/number format/fixed zerofill
		},
		yticklabel style = {font=\tiny},
		every axis label/.style={font=\tiny},
		title style={font=\footnotesize},
		grid=major,
		grid style={dashed, gray!30}
	]
	\addplot[line width=0.4mm, mark=*, color=violet!80!black]
		plot coordinates {
			(2, 81.39) (4, 81.25) (6, 81.30) (8, 81.38) (10, 81.39)
		};
	\end{axis}
	\end{tikzpicture}
}
\end{small}
\vspace{-2ex}
\caption{Varying the number \(K_{\text{nbr}}\) of neighboring clusters.}
\label{fig:hyperparameter_analysis_unique_style_Kneighbor}
\vspace{-2ex}
\end{figure}
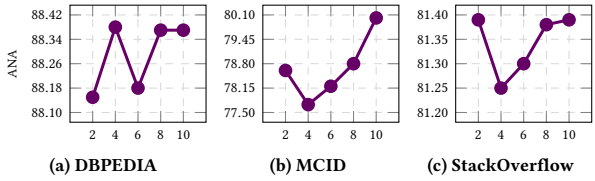

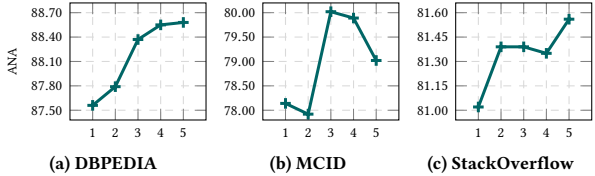
\begin{figure}[!t]
\centering
\begin{small}
\subfloat[DBPEDIA]{
	\begin{tikzpicture}[scale=1]
	\begin{axis}[
		height=\columnwidth/2.7,
		width=\columnwidth/2.5,
		ylabel={ANA},
		xmin=0, xmax=6,
		ymin=87.38, ymax=88.82,
		xtick={1, 2, 3, 4, 5},
		ytick={87.50, 87.80, 88.10, 88.40, 88.70},
		yticklabel style={
			font=\scriptsize,
			/pgf/number format/fixed,
			/pgf/number format/precision=2,
			/pgf/number format/fixed zerofill
		},
		yticklabel style = {font=\tiny},
		every axis label/.style={font=\tiny},
		title style={font=\footnotesize},
		grid=major,
		grid style={dashed, gray!30}
	]
	\addplot[line width=0.4mm, mark=+, color=teal!80!black]
		plot coordinates {
			(1, 87.56) (2, 87.79) (3, 88.37) (4, 88.55) (5, 88.58)
		};
	\end{axis}
	\end{tikzpicture}
}
\subfloat[MCID]{
	\begin{tikzpicture}[scale=1]
	\begin{axis}[
		height=\columnwidth/2.7,
		width=\columnwidth/2.5,
		xmin=0, xmax=6,
		ymin=77.8, ymax=80.2,
		xtick={1, 2, 3, 4, 5},
		ytick={78.00, 78.50, 79.00, 79.50, 80.00},
		yticklabel style={
			font=\scriptsize,
			/pgf/number format/fixed,
			/pgf/number format/precision=2,
			/pgf/number format/fixed zerofill
		},
		yticklabel style = {font=\tiny},
		every axis label/.style={font=\tiny},
		title style={font=\footnotesize},
		grid=major,
		grid style={dashed, gray!30}
	]
	\addplot[line width=0.4mm, mark=+, color=teal!80!black]
		plot coordinates {
			(1, 78.14) (2, 77.92) (3, 80.02) (4, 79.89) (5, 79.02)
		};
	\end{axis}
	\end{tikzpicture}
}
\subfloat[StackOverflow]{
	\begin{tikzpicture}[scale=1]
	\begin{axis}[
		height=\columnwidth/2.7,
		width=\columnwidth/2.5,
		xmin=0, xmax=6,
		ymin=80.94, ymax=81.66,
		xtick={1, 2, 3, 4, 5},
		ytick={81.00, 81.15, 81.30, 81.45, 81.60},
		yticklabel style={
			font=\scriptsize,
			/pgf/number format/fixed,
			/pgf/number format/precision=2,
			/pgf/number format/fixed zerofill
		},
		yticklabel style = {font=\tiny},
		every axis label/.style={font=\tiny},
		title style={font=\footnotesize},
		grid=major,
		grid style={dashed, gray!30}
	]
	\addplot[line width=0.4mm, mark=+, color=teal!80!black]
		plot coordinates {
			(1, 81.02) (2, 81.39) (3, 81.39) (4, 81.35) (5, 81.56)
		};
	\end{axis}
	\end{tikzpicture}
}
\end{small}
\vspace{-2ex}
\caption{Varying the number \(T\) of \algo{} iterations.}
\label{fig:hyperparameter_analysis_unique_style_tmax}
\vspace{-2ex}
\end{figure}

\subsection{Ablation Study}
To validate the effectiveness of the core components of \algo{}, we conduct a series of ablation studies.
As shown in Table~\ref{tab:ablation}, we analyze the influence of removing key components from \algo{} in both unsupervised and semi-supervised settings. In the unsupervised setting, removing either DCS or HSR leads to a noticeable drop in performance across all tested datasets. For example, on DBPedia, removing DCS decreases the NMI by 1.07\%, while removing HSR also degrades performance, confirming that both components are crucial for discovering high-quality clusters.

The same trend holds in the semi-supervised setting. Disabling DCS or HSR consistently lowers the NMI, ARI, and ACC scores. We also study the impact of removing the semi-supervised components: seeding and SML. The results show that both components contribute positively to the final performance. For instance, on M-CID, removing SML causes the ARI to drop from 73.20\% to 71.33\%. These findings underscore that each component in \algo{} plays an integral and synergistic role in its overall effectiveness.

\subsection{Parameter Analysis}

Let ANA be the average of ACC, NMI, and ARI. We conduct experiments to analyze the performance of \algo{} against strong competitors under different NID settings and the sensitivity of \algo{} to its key hyperparameters.

\begin{figure}[!t]
\centering
\begin{small}

\begin{tikzpicture}
    \begin{customlegend}[
        legend columns=3,
        legend entries={\algo, \LANID, \USNID},
        legend style={
            at={(0.5, 1.25)},
            anchor=north,
            draw=none,
            font=\footnotesize,
            column sep=0.2cm
        }
    ]
    \addlegendimage{line width=0.6mm, mark=*, color=myblue2}
    \addlegendimage{line width=0.6mm, mark=triangle, color=mycyan}
    \addlegendimage{line width=0.6mm, mark=square, color=myred_new2}
    \end{customlegend}
\end{tikzpicture}
\\[-\lineskip]
\vspace{-3ex}

\subfloat[\em DBPEDIA]{
\begin{tikzpicture}[scale=1, every mark/.append style={mark size=2pt}]
    \begin{axis}[
        height=\columnwidth/2.7,
        width=\columnwidth/2.5,
        every axis label/.style={font=\footnotesize},
        title style={font=\footnotesize},
        yticklabel style={
            font=\scriptsize,
            /pgf/number format/fixed,
            /pgf/number format/precision=2,
            /pgf/number format/fixed zerofill
        },
        yticklabel style = {font=\tiny},
        grid=major,
        grid style={dashed, gray!30},
        ylabel={ANA},
        xmin=0.2, xmax=0.8,
        xtick={0.25, 0.50, 0.75},
        xticklabel style = {font=\scriptsize},
        xticklabels={25\%, 50\%, 75\%},
        ymin=76.8, ymax=91.2,
        ytick={78.00, 81.00, 84.00, 87.00, 90.00},
    ]
    \addplot[line width=0.4mm, mark=*, color=myblue2]
        plot coordinates { (0.25, 78.88) (0.50, 80.76) (0.75, 88.96) };
    \addplot[line width=0.4mm, mark=triangle, color=mycyan]
        plot coordinates { (0.25, 78.52) (0.50, 80.37) (0.75, 81.32) };
    \addplot[line width=0.4mm, mark=square, color=myred_new2]
        plot coordinates { (0.25, 78.31) (0.50, 80.00) (0.75, 82.83) };
    \end{axis}
\end{tikzpicture}
}
\subfloat[\em MCID]{
\begin{tikzpicture}[scale=1, every mark/.append style={mark size=2pt}]
    \begin{axis}[
        height=\columnwidth/2.7,
        width=\columnwidth/2.5,
        every axis label/.style={font=\footnotesize},
        title style={font=\footnotesize},
        yticklabel style={
            font=\scriptsize,
            /pgf/number format/fixed,
            /pgf/number format/precision=2,
            /pgf/number format/fixed zerofill
        },
        yticklabel style = {font=\tiny},
        grid=major,
        grid style={dashed, gray!30},
        xmin=0.2, xmax=0.8,
        xtick={0.25, 0.50, 0.75},
        xticklabel style = {font=\scriptsize},
        xticklabels={25\%, 50\%, 75\%},
        ymin=62.8, ymax=83.2, 
        ytick={64.00, 68.50, 73.00, 77.50, 82.00}, 
    ]
    \addplot[line width=0.4mm, mark=*, color=myblue2]
        plot coordinates { (0.25, 73.69) (0.50, 77.23) (0.75, 80.57) };
    \addplot[line width=0.4mm, mark=triangle, color=mycyan]
        plot coordinates { (0.25, 68.46) (0.50, 72.68) (0.75, 77.82) };
    \addplot[line width=0.4mm, mark=square, color=myred_new2]
        plot coordinates { (0.25, 64.09) (0.50, 64.88) (0.75, 74.80) };
    \end{axis}
\end{tikzpicture}
}
\subfloat[\em StackOverflow]{
\begin{tikzpicture}[scale=1, every mark/.append style={mark size=2pt}]
    \begin{axis}[
        height=\columnwidth/2.7,
        width=\columnwidth/2.5,
        every axis label/.style={font=\footnotesize},
        title style={font=\footnotesize},
        yticklabel style={
            font=\scriptsize,
            /pgf/number format/fixed,
            /pgf/number format/precision=2,
            /pgf/number format/fixed zerofill
        },
        yticklabel style = {font=\tiny},
        grid=major,
        grid style={dashed, gray!30},
        xmin=0.2, xmax=0.8,
        xtick={0.25, 0.50, 0.75},
        xticklabel style = {font=\scriptsize},
        xticklabels={25\%, 50\%, 75\%},
        ymin=70.8, ymax=83.2,
        ytick={72.00, 74.50, 77.00, 79.50, 82.00},
    ]
    \addplot[line width=0.4mm, mark=*, color=myblue2]
        plot coordinates { (0.25, 75.88) (0.50, 78.97) (0.75, 81.56) };
    \addplot[line width=0.4mm, mark=triangle, color=mycyan]
        plot coordinates { (0.25, 74.86) (0.50, 78.53) (0.75, 72.48) };
    \addplot[line width=0.4mm, mark=square, color=myred_new2]
        plot coordinates { (0.25, 72.04) (0.50, 77.49) (0.75, 80.12) };
    \end{axis}
\end{tikzpicture}
}
\end{small}
\vspace{-2ex}
\caption{Varying KCR.}
\label{fig:kcr_comparison}
\vspace{-2ex}
\end{figure}
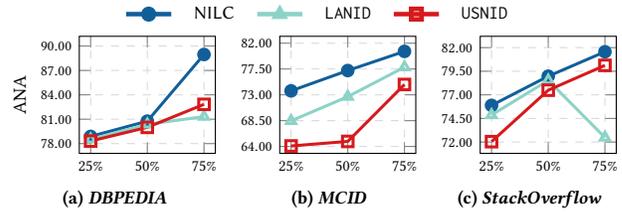

\stitle{Known Class Ratio} Fig.~\ref{fig:kcr_comparison} illustrates the performance of our method compared to strong baselines (\LANID, \USNID) as the Known Class Ratio (KCR) varies in \{25\%, 50\%, 75\%\}. Across different datasets (DBPedia, MCID, and StackOverflow), our method consistently outperforms the others at every KCR level. Notably, the performance of all methods generally improves with a higher KCR, which is expected as more labeled data provides better supervision. However, the performance gap between our method and the competitors remains significant, highlighting the robustness of \algo{} even in low-resource settings.

\stitle{Hyperparameter Sensitivity} We analyze the impact of the clustering cost weights $\alpha$, $\beta$, and $\gamma$ on DBPEDIA, MCID, and StackOverflow, with results shown in Figs.~\ref{fig:hyperparameter_alpha}, ~\ref{fig:hyperparameter_beta}, and ~\ref{fig:hyperparameter_gamma}. We can observe that \algo{}'s performance remains stable across a wide range of values for each hyperparameter. For example, varying $\alpha$ in \{0.1, 0.3, 0.5, 0.7, 0.9\} on StackOverflow results in only minor fluctuations in ANA, which stays within a tight range of 80.86 to 81.24. This indicates that our method is not overly sensitive to the precise tuning of these weights, making it practical for real-world applications.

\stitle{In-Context Learning Parameters} We examine the influence of key in-context learning (ICL) parameters. For selected hard samples, as shown in Fig.~\ref{fig:hyperparameter_analysis_unique_style_O}, it shows that a moderate number in \{10, 15\} often yields the best results on MCID, while performance on other datasets is less sensitive. Fig.~\ref{fig:hyperparameter_analysis_unique_style_Kneighbor} shows the impact of the number of neighboring clusters used for context. Performance is generally robust, with the tendency that higher values deliver stronger results. Finally, Fig.~\ref{fig:hyperparameter_analysis_unique_style_tmax} indicates that the model benefits from multiple LLM-driven \algo{} iterations, with performance generally improving or stabilizing after 3 iterations. This suggests that a few rounds of refinement are sufficient to achieve significant gains.

\stitle{Pre-trained Text Encoders} We investigate the impact of different $\textsf{PTE}(\cdot)$ on \algo{}'s performance. As shown in Table~\ref{tab:performance_encoders}, we replace \USNID with four models: \SentenceBERT~\cite{reimers2019sentence}, \Instructor~\cite{su2022one}, \MTPCLNN, and \LatentEM. The results clearly demonstrate that \algo{} consistently enhances the performance of all encoders across DBPedia, M-CID, and StackOverflow. For instance, when applied to \Instructor on DBPedia, \algo{} improves the NMI from 70.79\% to 77.70\%, ARI from 57.65\% to 64.66\%, and ACC from 71.86\% to 73.43\%. This underscores the robustness and versatility of \algo{}, as it is not dependent on a specific encoder but can effectively augment various text representation models to achieve better clustering outcomes.

\stitle{Large Language Models} We evaluate the performance of \algo{} with different LLMs, including \GPTMini, \GPT, \Qwen, \Gemini, and \Deepseek. As shown in Fig.~\ref{fig:llm_comparison}, all LLMs achieve competitive results, indicating that our framework is robust and not overly sensitive to the choice of a specific LLM. This flexibility allows users to choose an LLM that best fits computational and financial constraints without a significant drop in performance, highlighting the practical applicability of \algo{}.

\definecolor{lightblue}{RGB}{173, 216, 230}
\definecolor{myorange}{RGB}{255, 165, 0}
\definecolor{mypink}{RGB}{255, 182, 193}
\definecolor{mycyan}{RGB}{0, 255, 255}
\definecolor{mygreen}{RGB}{144, 238, 144}

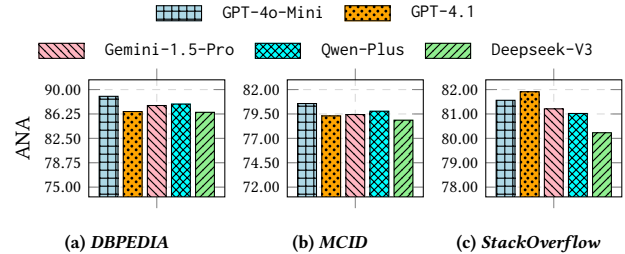
\begin{figure}[!t]
\centering
\begin{small}

\begin{tikzpicture}
    \begin{customlegend}[
        legend entries={{\texttt{GPT-4o-Mini}}, {\texttt{GPT-4.1}}},
        legend columns=2,
        area legend,
        legend style={
            at={(0.5, 1.6)},
            anchor=north,
            draw=none,
            font=\footnotesize,
            column sep=0.2cm
        }
    ]
    \addlegendimage{preaction={fill, lightblue}, pattern=grid}
    \addlegendimage{preaction={fill, myorange}, pattern=crosshatch dots}
    \end{customlegend}

    \begin{customlegend}[
        legend entries={{\texttt{Gemini-1.5-Pro}}, {\texttt{Qwen-Plus}}, {\texttt{Deepseek-V3}}},
        legend columns=3,
        area legend,
        legend style={
            at={(0.5, 1.1)},
            anchor=north,
            draw=none,
            font=\footnotesize,
            column sep=0.2cm
        }
    ]
    \addlegendimage{preaction={fill, mypink}, pattern=north west lines}
    \addlegendimage{preaction={fill, mycyan}, pattern=crosshatch}
    \addlegendimage{preaction={fill, mygreen}, pattern=north east lines}
    \end{customlegend}
\end{tikzpicture}
\\[-\lineskip]
\vspace{-3ex}

\subfloat[\em DBPEDIA]{
\begin{tikzpicture}[scale=1]
    \begin{axis}[
        height=\columnwidth/2.7,
        width=\columnwidth/2.5,
        ybar,
        bar width=0.25cm,
        enlarge x limits=0.5,
        symbolic x coords={DBPEDIA},
        xtick=data,
        xticklabels={},
        ylabel={ANA},
        yticklabel style={             font=\scriptsize,             /pgf/number format/fixed,             /pgf/number format/precision=2,             /pgf/number format/fixed zerofill         },
        ymin=73.5, ymax=91.5,
        ytick={75, 78.75, 82.5, 86.25, 90},
        yticklabel style={             font=\scriptsize,             /pgf/number format/fixed,             /pgf/number format/precision=2,             /pgf/number format/fixed zerofill         },
        grid=major,
        grid style={dashed, gray!30},
    ]
    \addplot [preaction={fill, lightblue}, pattern={grid}] coordinates {(DBPEDIA, 88.96)};
    \addplot [preaction={fill, myorange}, pattern={crosshatch dots}] coordinates {(DBPEDIA, 86.62)};
    \addplot [preaction={fill, mypink}, pattern=north west lines] coordinates {(DBPEDIA, 87.55)};
    \addplot [preaction={fill, mycyan}, pattern=crosshatch] coordinates {(DBPEDIA, 87.77)};
    \addplot [preaction={fill, mygreen}, pattern={north east lines}] coordinates {(DBPEDIA, 86.50)};
    \end{axis}
\end{tikzpicture}
}
\subfloat[\em MCID]{
\begin{tikzpicture}[scale=1]
    \begin{axis}[
        height=\columnwidth/2.7,
        width=\columnwidth/2.5,
        ybar,
        bar width=0.25cm,
        enlarge x limits=0.5,
        symbolic x coords={MCID},
        xtick=data,
        xticklabels={},
        ylabel={},
        ymin=71, ymax=83,
        ytick={72, 74.5, 77, 79.5, 82},
        yticklabel style={             font=\scriptsize,             /pgf/number format/fixed,             /pgf/number format/precision=2,             /pgf/number format/fixed zerofill         },
        grid=major,
        grid style={dashed, gray!30},
    ]
    \addplot [preaction={fill, lightblue}, pattern={grid}] coordinates {(MCID, 80.57)};
    \addplot [preaction={fill, myorange}, pattern={crosshatch dots}] coordinates {(MCID, 79.30)};
    \addplot [preaction={fill, mypink}, pattern=north west lines] coordinates {(MCID, 79.44)};
    \addplot [preaction={fill, mycyan}, pattern=crosshatch] coordinates {(MCID, 79.78)};
    \addplot [preaction={fill, mygreen}, pattern={north east lines}] coordinates {(MCID, 78.86)};
    \end{axis}
\end{tikzpicture}
}
\subfloat[\em StackOverflow]{
\begin{tikzpicture}[scale=1]
    \begin{axis}[
        height=\columnwidth/2.7,
        width=\columnwidth/2.5,
        ybar,
        bar width=0.25cm,
        enlarge x limits=0.5,
        symbolic x coords={StackOverflow},
        xtick=data,
        xticklabels={},
        ylabel={},
        ymin=77.6, ymax=82.4,
        ytick={78, 79, 80, 81, 82},
        yticklabel style={             font=\scriptsize,             /pgf/number format/fixed,             /pgf/number format/precision=2,             /pgf/number format/fixed zerofill         },
        grid=major,
        grid style={dashed, gray!30},
    ]
    \addplot [preaction={fill, lightblue}, pattern={grid}] coordinates {(StackOverflow, 81.56)};
    \addplot [preaction={fill, myorange}, pattern={crosshatch dots}] coordinates {(StackOverflow, 81.91)};
    \addplot [preaction={fill, mypink}, pattern=north west lines] coordinates {(StackOverflow, 81.21)};
    \addplot [preaction={fill, mycyan}, pattern=crosshatch] coordinates {(StackOverflow, 81.02)};
    \addplot [preaction={fill, mygreen}, pattern={north east lines}] coordinates {(StackOverflow, 80.23)};
    \end{axis}
\end{tikzpicture}
}
\end{small}
\vspace{-2ex}
\caption{Varying LLMs.}
\label{fig:llm_comparison}
\vspace{-2ex}
\end{figure}

\section{Conclusion}
In this paper, we propose \algo{}, a framework for New Intent Discovery that synergizes embedding-based clustering and Large Language Models. Our method iteratively refines cluster assignments and text embeddings, featuring the dual centroid scheme (Euclidean and LLM-semantic) and an integrated hard sample refinement mechanism. We also demonstrate how to inject semi-supervised signals through seeding and soft must-links. Experiments on six benchmark datasets show that \algo{} consistently achieves state-of-the-art performance in both unsupervised and semi-supervised settings, validating our integrated, LLM-assisted approach.

\begin{acks}
This work is partially supported by the National Natural Science Foundation of China (No. 62302414), the Hong Kong RGC ECS grant (No. 22202623) and YCRG (No. C2003-23Y), the Huawei Gift Fund, and Guangdong and Hong Kong Universities ``1+1+1'' Joint Research Collaboration Scheme, project No.: 2025A0505000002.
\end{acks}

\balance

\section*{Ethical Considerations}
The direct negative societal impacts of this research—specifically with respect to fairness, privacy, and security—are minimal. Nonetheless, as with other NID solutions, erroneous results by the method may affect system functionality. Although the algorithm's effectiveness has been extensively validated through experiments, occasional inaccuracies, particularly when processing noisy data, are still possible. To mitigate these risks, it is recommended to enhance data quality through rigorous data cleaning and preprocessing prior to method deployment.

\bibliographystyle{ACM-Reference-Format}
\bibliography{sample-base}

\appendix
\section{Notation}
Table~\ref{tab:notations} provides a summary of the key notations used throughout this paper.

\begin{table}[h]
\centering
\caption{Summary of notations.}
\vspace{-2ex}
\label{tab:notations}
\small
\resizebox{\columnwidth}{!}{%
\begin{tabular}{ll}
\toprule
\textbf{Symbol} & \textbf{Description} \\
\midrule
$\Y, \Y_k, \Y_u$ & Sets of all, known, and unknown intents. \\
$K, M$ & Total and known intent counts. \\
$\D_l, \D_u$ & Sets of labeled and unlabeled utterances. \\
$\D_{train}, \D_{test}$ & Training and testing sets. \\
$X, x_i$ & Set of utterances and the $i$-th utterance. \\
$\X, \xvec_i$ & Set of utterance embeddings and the embedding for $x_i$. \\
$y_i$ & The $i$-th utterance's intent. \\
$N, d$ & Number of utterances and embedding dimension. \\
$T, t$ & Total iterations and current iteration index. \\
$\textsf{PTE}(\cdot)$ & Pre-trained Text Encoder. \\
$\textsf{LLM}(\cdot)$ & Large Language Model for generation and refinement. \\
\midrule
$p_{\text{smry}}$ & The prompt template for summary generation. \\
$\C_k, \boldsymbol{\mu}_k$ & The $k$-th cluster and its Euclidean centroid. \\
$\boldsymbol{\theta}_k, s_k$ & The semantic centroid and textual summary for cluster $\C_k$. \\
$\boldsymbol{\theta}_{\text{nbr}(k)}$ & The nearest neighboring semantic centroid to $\boldsymbol{\theta}_k$. \\
$\Sset_k$ & Set of representative exemplars from cluster $\C_k$. \\
$f(\cdot)$ & The clustering cost function. \\
$\textsf{cos}(\cdot)$ & Cosine similarity between two utterance embeddings. \\
$\mathcal{L}^{\text{ED}}$ & The Euclidean distance cost. \\
$\mathcal{L}^{\text{SC}}, \alpha$ & The semantic cohesion cost and its weight. \\
$\mathcal{L}^{\text{SS}}, \beta$ & The semantic separation cost and its weight. \\
\midrule
$p_{\text{ref}}$ & The prompt template for utterance refinement. \\
$\Hset$ & Set of identified hard samples for refinement. \\
$H(\cdot), \delta$ & Shannon entropy function and the number of hard samples. \\
$K_{\text{nbr}}$ & The number of neighboring clusters for HSR context. \\
$\tilde{x}_i, \tilde{\xvec}_i$ & Refined utterance and its new embedding. \\
$s_h, \Sset_h$ & Summary and exemplars of a sample's home cluster. \\
\midrule
$\{\boldsymbol{\mu}_k^{0}\}_{k=1}^K$ & The initial Euclidean centroids. \\
$\{\boldsymbol{\mu}^*_j\}_{j=1}^M$ & Seed centroids from labeled data $\D_l$. \\
$\mathcal{L}^{\text{SP}}, \gamma$ & The supervised cost and its weight. \\
$\textsf{Mean}(\cdot)$ & The mean of a set of embeddings. \\
\bottomrule
\end{tabular}%
}
\end{table}

\section{Baseline Repositories}
Table~\ref{tab:baselines} lists the public code repositories used for the baseline methods in our experiments.

\begin{table}[H]
\centering
\caption{Code repositories for baselines.}
\label{tab:baselines}
\vspace{-2ex}
\small
\resizebox{\columnwidth}{!}{%
\begin{tabular}{ll}
\toprule
\textbf{Baseline} & \textbf{Code Repository} \\

\midrule
\SAE/\DEC & \url{https://github.com/piiswrong/dec} \\
\DCN & \url{https://github.com/boyangumn/DCN} \\
\CC & \url{https://github.com/XLearning-SCU/2021-AAAI-CC} \\
\SCCL & \url{https://github.com/amazon-science/sccl} \\
\KCL/\MCL & \url{https://github.com/GT-RIPL/L2C} \\
\DTC & \url{https://github.com/k-han/DTC} \\
\CDAC & \url{https://github.com/thuiar/CDAC-plus} \\
\GCD & \url{https://github.com/sgvaze/generalized-category-discovery} \\
\DeepAligned & \url{https://github.com/HanleiZhang/DeepAligned-Clustering} \\
\MTPCLNN & \url{https://github.com/fanolabs/NID_ACLARR2022} \\
\LatentEM & \url{https://github.com/zyh190507/Probabilistic-discovery-new-intents} \\
\USNID & \url{https://github.com/thuiar/TEXTOIR} \\
\SDC & \url{https://github.com/Lackel/SDC} \\
\LANID & \url{https://github.com/floatSDSDS/LANID} \\

\bottomrule
\end{tabular}%
}
\end{table}

\section{Hyperparameter Settings}
\label{sec:param-set}
Table~\ref{tab:hyperparameters_detailed} details the hyperparameter configurations. Across all settings, we employ \USNID as $\textsf{PTE}(\cdot)$ and \GPTMini as $\textsf{LLM}(\cdot)$. We use fixed parameters for DCS and HSR: $|\Sset_k| = 10$, $\delta = 10$, and $K_{\text{nbr}} = 10$. For brevity, these constant values are omitted from the main hyperparameter table.

We can observe that MMR proves to be the most effective choice in the majority of cases. Notably, for the semi-supervised experiments on the BANKING and CLINC datasets, we opt for a similarity-based mapping strategy instead of one reliant on the LLM. This is because these datasets feature a large number of known intents (over a hundred). The LLM struggles to fully comprehend and accurately map such a wide range of intents while strictly adhering to the required output format. Consequently, a direct similarity-based mapping provides a more robust and effective solution in high-cardinality scenarios.

\begin{table}[h]
\centering
\caption{Hyperparameter settings for \algo{}.}
\label{tab:hyperparameters_detailed}
\vspace{-2ex}
\resizebox{\columnwidth}{!}{%
\begin{tabular}{l l c c c c c c}
\toprule
\textbf{Setting} & \textbf{Dataset} & \textbf{Selection Strategy} & \textbf{$T$} & \textbf{$\alpha$} & \textbf{$\beta$} & \textbf{$\gamma$} & \textbf{Mapping Strategy} \\
\midrule
\multirow{6}{*}{\STAB{\rotatebox[origin=c]{90}{Unsupervised}}} 
& BANKING & MMR & 3 & 0.5 & 0.5 & -- & -- \\
& CLINC & MMR & 3 & 0.5 & 0.3 & -- & -- \\
& DBPedia & MMR & 3 & 0.9 & 0.5 & -- & -- \\
& M-CID & NN & 2 & 0.3 & 0.3 & -- & -- \\
& SNIPS & MMR & 3 & 0.3 & 0.5 & -- & -- \\
& StackOverflow & MMR & 3 & 0.9 & 0.1 & -- & -- \\
\midrule
\multirow{6}{*}{\STAB{\rotatebox[origin=c]{90}{Semi-Supervised}}} 
& BANKING & MMR & 3 & 0.9 & 0.3 & 0.5 & Similarity-based \\
& CLINC & MMR & 3 & 0.5 & 0.5 & 0.5 & Similarity-based \\
& DBPedia & MAD & 3 & 0.5 & 0.5 & 0.5 & LLM-based \\
& M-CID & NN & 3 & 0.3 & 0.1 & 0.5 & LLM-based \\
& SNIPS & MMR & 3 & 0.5 & 0.5 & 0.5 & LLM-based \\
& StackOverflow & MMR & 3 & 0.9 & 0.7 & 0.1 & LLM-based \\
\bottomrule
\end{tabular}%
}
\end{table}

\section{Selection Strategies for $\Sset_k$}\label{sec:selection-strategy}

\stitle{$K$-\texttt{Means++}}
This strategy adapts the seeding procedure of $K$-\texttt{Means++} to select a geometrically diverse set of exemplars. The selection is iterative. Let $\Sset_k^{(i)}$ be the set of $i$ selected exemplars. The first exemplar, $\xvec_1$, is chosen uniformly at random from $\C_k$ to form $\Sset_k^{(1)}$. For $i = 2, \dots, |\Sset_k|$, each subsequent exemplar $\xvec_i$ is chosen from the remaining embeddings $\C_k \setminus \Sset_{k}^{(i-1)}$ with a probability proportional $G(\xvec_i)$ to its minimum squared Euclidean distance to the set of already-selected exemplars $\Sset_{k}^{(i-1)}$:
\begin{equation}
    G(\xvec_i) = \frac{\min_{\xvec_s \in \Sset_{k}^{(i-1)}} ||\xvec_i - \xvec_s||^2}{\sum_{\xvec_j \in \C_k \setminus \Sset_{k}^{(i-1)}} \min_{\xvec_s \in \Sset_{k}^{(i-1)}} ||\xvec_j - \xvec_s||^2}
\end{equation}
This method is designed to maximize the diversity of $\Sset_k$, ensuring broad coverage of the cluster's semantic space.

\stitle{{Mean Average Distance} ({MAD})}
The MAD strategy identifies exemplars from the cluster's periphery by selecting those that are, on average, most dissimilar from other members of the cluster. We select the set $\Sset_k$ by maximizing the mean distance:
\begin{equation}
    \Sset_k = \argmax{\Sset \subset \C_k, |\Sset|=|\Sset_k|} \sum_{\xvec_i \in \Sset} \frac{1}{|\C_k|-1} \sum_{\xvec_j \in \C_k, j \neq i} ||\xvec_i - \xvec_j||
\end{equation}
The theoretical justification is that these boundary points are crucial for defining the cluster's extent and improving its separation from neighboring clusters.

\stitle{{Maximal Marginal Relevance} ({MMR})}
MMR provides a formal framework for balancing relevance to the cluster's central theme with the diversity of the selected exemplars. After an initial exemplar is chosen based on maximum similarity to the geometric centroid $\boldsymbol{\mu}_k$, subsequent exemplars are selected iteratively to maximize the following objective function:
\begin{equation}
    \argmax{\xvec_j \in \C_k \setminus \Sset_k} \left[ \text{cos}(\xvec_j, \boldsymbol{\mu}_k) - \max_{\xvec_s \in \Sset_k} \text{cos}(\xvec_j, \xvec_s) \right]
\end{equation}
This ensures that $\Sset_k$ is composed of exemplars that are both highly representative and non-redundant.

\stitle{Nearest Neighbors ({NN})}
This strategy selects the most central and prototypical instances of the cluster. The centrality $C(\xvec_i)$ of an embedding is defined as its cumulative similarity to all other embeddings within the cluster:
\begin{equation}
    C(\xvec_i) = \sum_{\xvec_j \in \C_k, j \neq i} \text{cos}(\xvec_i, \xvec_j)
\end{equation}
The set $\Sset_k$ is formed by the utterances corresponding to the embeddings with the highest centrality scores. The premise is that the most central points are the most faithful representatives of the underlying intent.

\begin{figure}[!t]
    \centering
    \includegraphics[width=0.98\columnwidth]{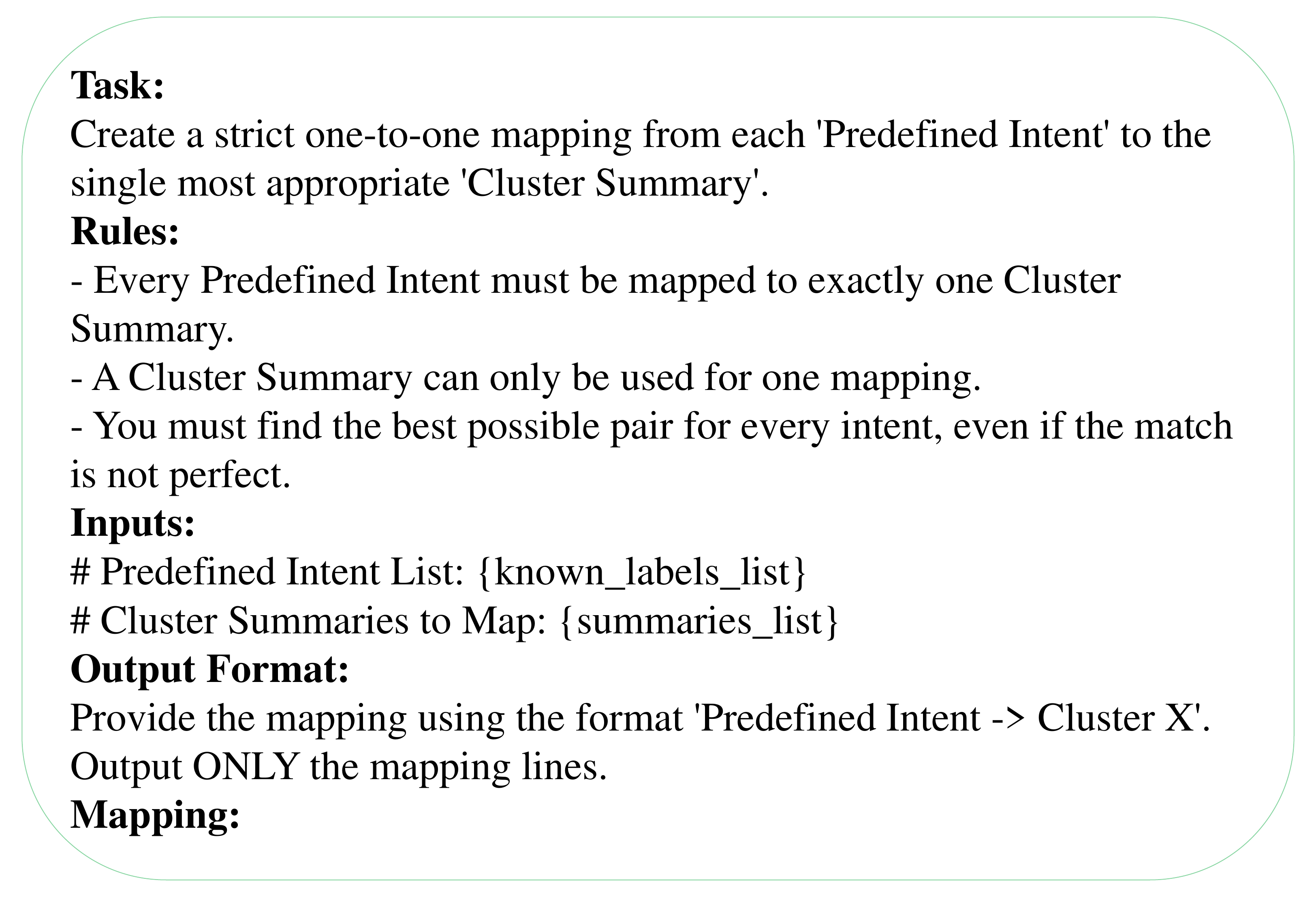} 
    \vspace{-2ex}
    \caption{Prompt template for LLM-based mappings.}
    \label{fig:nid-mapping-template}
\vspace{0ex}
\end{figure}

\section{Mapping Strategies for $\pi^t$}\label{sec:map-strategy}

\stitle{Embedding-based Mapping} This approach matches the known seed centroids $\{\boldsymbol{\mu}^*_j\}_{j=1}^M$ to the current semantic centroids $\{\boldsymbol{\theta}_k^{t}\}_{k=1}^K$ by solving the assignment problem that minimizes cosine dissimilarity, using the Hungarian algorithm:
\begin{equation}
\min_{\pi^{t}} \sum_{m=1}^{M} \left(1 - \text{cos}(\boldsymbol{\mu}^*_j, \boldsymbol{\theta}_{\pi^{t}(j)}^{t})\right)
\end{equation}

\stitle{LLM-based Mapping} This strategy employs the LLM to perform a direct semantic mapping between the known intent labels $\Y_k$ and the generated cluster summaries $\{s_k^{t}\}_{k=1}^K$. As detailed in Fig.~\ref{fig:nid-mapping-template}, the prompt $p_{\text{map}}$ is designed to constrain the LLM to behave like an optimal assignment algorithm to create a strict one-to-one mapping by enforcing rules that require every intent to be matched with a unique cluster summary. This process compels the LLM to find the best possible pairing for each intent:
\begin{equation}
\pi^{t} = \textsf{LLM}\left(p_{\text{map}}, \Y_k, \{s_k^{t}\}_{k=1}^K\right)
\end{equation}

\section{Case Studies}
\label{sec:case_studies}

\subsection{Evolution of Semantic Centroids}
\label{sec:case-study-evolution}
To understand how \algo{} iteratively refines the semantic understanding of each cluster, we can dive into the evolution of the LLM-generated cluster summaries, which act as semantic centroids. Table~\ref{tab:case_study_evolution} presents a case study from M-CID, tracking the summary $s_{10}$ for $\C_{10}$ over 3 iterations.

Initially, the cluster is summarized with a broad question about general cleaning practices. As the cluster assignments and embeddings are refined, the summary evolves to become more specific and detailed. By Iteration 2, it narrows its focus to the survivability of the virus on surfaces. Finally, in Iteration 3, the summary crystallizes into a precise and actionable utterance about the ``best practices and precautions'' for preventing surface transmission. This progressive refinement demonstrates how \algo{}'s iterative process enhances the coherence and specificity of the discovered intents.

\begin{table}[!t]
\centering
\caption{Evolution of $s_{10}$ for $\C_{10}$ on M-CID.}
\label{tab:case_study_evolution}
\vspace{-2ex}
\small
\resizebox{\columnwidth}{!}{%
\begin{tabular}{c|p{0.8\columnwidth}}
\toprule
\textbf{Iteration} & \textbf{Summary $s_{10}$ for Cluster $\C_{10}$} \\
\midrule
\textbf{1} & What cleaning and disinfecting practices are effective in preventing the spread of COVID-19 on surfaces? \\
\midrule
\textbf{2} & How long does the coronavirus survive on various surfaces and materials, and what cleaning practices are recommended? \\
\midrule
\textbf{3} & What are the best cleaning practices and precautions to prevent COVID-19 transmission from surfaces and packages? \\
\bottomrule
\end{tabular}%
}
\end{table}

\begin{table}[!t]
\centering
\caption{An example of the HSR process for an ambiguous utterance on StackOverflow.}
\vspace{-2ex}
\label{tab:case_study_refinement}
\small
\resizebox{\columnwidth}{!}{%
\begin{tabular}{p{0.25\columnwidth}|p{0.7\columnwidth}}
\toprule
\textbf{Component} & \textbf{Content} \\
\midrule
\textbf{Hard Sample ($x_i$)} & Confusion regarding laziness \\
\midrule
\textbf{Assigned Cluster ($\C_4$)} & \textbf{Summary ($s_4$):} What are the various techniques and best practices for effectively using LINQ to query and manipulate data, including handling distinct values, dynamic queries, joins, and return types? \\
\midrule
\textbf{True Cluster ($\C_6$)} & \textbf{Summary ($s_6$):} What are some common challenges and best practices when working with Haskell, including syntax, error handling, and functional constructs? \\
\midrule
\textbf{LLM Task} & Given the context, analyze the best fit for the utterance and rewrite it to be an unambiguous exemplar of that theme. \\
\midrule
\textbf{Refined Utterance ($\tilde{x}_i$)} & Understanding laziness in functional programming languages like Haskell \\
\bottomrule
\end{tabular}%
}
\end{table}

\begin{table}[!t]
\centering
\caption{Comparison of mapping strategies on DBPedia.}
\label{tab:case_study_mapping_detailed}
\vspace{-2ex}
\resizebox{\columnwidth}{!}{%
\begin{tabular}{l|l|l}
\toprule
\textbf{Mapping Strategy} & \textbf{Summary $s_k$ for Cluster $\C_k$} & \textbf{Mapped Known Intent} \\
\midrule
\multirow{10}{*}{Similarity-based} 
& $\C_0$: Books and Publications & WrittenWork \\
& $\C_1$: Notable Individuals & OfficeHolder \\
& $\C_2$: Plant Species & Plant \\
& $\C_3$: Organisms & Animal \\
& $\C_4$: Historic and Cultural Institutions & Building \\
& $\C_6$: Historical Vehicles and Vessels & MeanOfTransportation \\
& $\C_8$: Diverse Companies and Organizations & Company \\
& $\C_9$: Films & Film \\
& $\C_{10}$: Geographical Features & NaturalPlace \\
& \textbf{$\C_{13}$: Professional Athlete} & \textbf{Artist} \\
\midrule
\multirow{10}{*}{LLM-based} 
& $\C_0$: Literary Works & WrittenWork \\
& $\C_1$: Notable Individuals & OfficeHolder \\
& $\C_2$: Plant Species & Plant \\
& $\C_3$: Organisms & Animal \\
& $\C_4$: Historic and Cultural Institutions & Building \\
& \textbf{$\C_5$: Music Albums and Compilations} & \textbf{Artist} \\
& $\C_6$: Historical Vehicles and Vessels & MeanOfTransportation \\
& $\C_8$: Corporations and Organizations & Company \\
& $\C_9$: Films & Film \\
& $\C_{10}$: Geographical Features & NaturalPlace \\
\bottomrule
\end{tabular}%
}
\end{table}

\subsection{Successful Refinement of Hard Samples}
\label{sec:case-study-hsr}

To illustrate the utility of HSR, we present a qualitative example on StackOverflow. HSR clarifies ambiguous utterances by leveraging the LLM's contextual understanding.

For instance, the utterance $x_i$ = ``Confusion regarding laziness'' is ambiguous. It was initially misclassified into a cluster about LINQ queries because ``laziness'' can relate to deferred execution, which is not the utterance's core intent. \algo{} identifies this high-uncertainty sample and provides the LLM with the context of its assigned and neighboring clusters, as detailed in Table~\ref{tab:case_study_refinement}. Note that the ``True Cluster'' is shown for illustrative purposes; the LLM only receives the ``home'' cluster and neighboring clusters as the context, not its ground-truth identity.

The LLM analyzes the competing intents and, recognizing that ``laziness'' is a core concept in Haskell, rewrites the utterance into a clear and specific question. The new embedding $\tilde{\xvec}_i$ for the refined utterance has a much lower clustering cost and is confidently reassigned to the correct Haskell-related cluster. This case study demonstrates how HSR actively corrects the data manifold, improving cluster cohesion and separation by resolving ambiguity.

\subsection{Comparison of Two Mapping Strategies}
\label{sec:case-study-mapping}
To showcase the superiority of our LLM-based mapping approach over the traditional similarity-based method, we present a detailed comparison of the mappings generated for DBPedia, as shown in Table~\ref{tab:case_study_mapping_detailed}.

While both methods successfully map many clusters, the similarity-based approach, which relies on cosine distance between centroids, makes a critical error. It incorrectly maps $\C_{13}$, summarized as ``Professional Athlete'', to the known intent ``Artist''. Although athletes can be metaphorically considered ``artists'' of sports, this is not the correct semantic relationship on DBPedia. This error underscores a fundamental limitation of relying purely on embedding similarity in a Euclidean space; such an approach is confined to geometric proximity and lacks the awareness of semantic context, possibly causing it to be misled by abstract or metaphorical connections that an LLM, with its richer world knowledge, can correctly disambiguate. 

In contrast, our LLM-based method leverages its world knowledge and reasoning capabilities. It correctly discerns that ``Professional Athlete'' does not fit the ``Artist'' category and instead makes the more semantically sound decision to map $\C_5$ (``Music Albums and Compilations'') to ``Artist''. This leads to a more effective injection of semi-supervised signals, a more accurate must-links constraint relationship, and, ultimately, a more accurate final clustering.

\begin{table}[!t]
\renewcommand{\arraystretch}{0.9}
\caption{Analysis of Semi-Supervised Mapping Strategies.}
\label{tab:mapping_strategy_comparison}
\vspace{-3ex}
\begin{center}
\small
\begin{tabular}{l|l|c|c|c}
\toprule
{Dataset} & {Mapping Strategy} & {NMI} & {ARI} & {ACC} \\
\midrule
\multirow{2}{*}{DBPedia}       & Similarity-based & 89.41 & 83.96 & 91.43 \\
                                         & LLM-based        & 89.36 & 84.19 & 91.57 \\
\midrule
\multirow{2}{*}{M-CID}         & Similarity-based & 81.49 & 70.56 & 83.09 \\
                                         & LLM-based        & 83.06 & 72.48 & 84.53 \\
\midrule
\multirow{2}{*}{StackOverflow} & Similarity-based & 80.13 & 75.89 & 86.85 \\
                                         & LLM-based        & 80.53 & 76.47 & 87.18 \\
\bottomrule
\end{tabular}
\end{center}
\vspace{0ex}
\end{table}

\begin{table}[!t]
\renewcommand{\arraystretch}{0.9}
\caption{Analysis of Representative Sampling Methods.}
\label{tab:rep_method_comparison}
\vspace{-3ex}
\begin{center}
\small
\begin{tabular}{l|l|c|c|c}
\toprule
{Dataset} & {Selection Strategy} & {NMI} & {ARI} & {ACC} \\
\midrule
\multirow{4}{*}{DBPedia} & MMR & 89.36 & 84.19 & 91.57 \\
 & MAD & 89.99 & 84.88 & 92.00 \\
 & NN & 88.90 & 83.81 & 91.43 \\
 & $K$-\texttt{Means++} & 89.36 & 84.03 & 91.43 \\
\midrule
\multirow{4}{*}{M-CID} & MMR & 83.06 & 72.48 & 84.53 \\
 & MAD & 81.83 & 70.64 & 83.09 \\
 & NN & 83.36 & 73.12 & 85.10 \\
 & K-Means++ & 82.62 & 71.66 & 83.67 \\
\midrule
\multirow{4}{*}{StackOverflow} & MMR & 80.53 & 76.47 & 87.18 \\
 & MAD & 80.19 & 75.86 & 86.77 \\
 & NN & 80.33 & 76.15 & 86.93 \\
 & $K$-\texttt{Means++} & 80.28 & 76.16 & 86.95 \\
\bottomrule
\end{tabular}
\end{center}
\vspace{0ex}
\end{table}

\section{Empirical Studies of Mapping and Sampling Strategies}
\label{sec:empiracal-study-mapping-sampling}
We further analyze the specific strategies used to represent clusters. Table~\ref{tab:mapping_strategy_comparison} compares the LLM-based mapping strategy against a traditional similarity-based approach for semi-supervised NID. The LLM-based strategy consistently outperforms the similarity-based one, particularly on the M-CID dataset, where it yields a 1.92\% improvement in ARI. This shows that LLMs can capture the semantic alignment between known intent labels and cluster summaries more effectively than simple embedding similarity. In Table~\ref{tab:rep_method_comparison}, we analyze different representative sampling methods for generating cluster summaries. While all methods perform well, the MAD, NN, and MMR strategies show slight advantages on DBPedia,  M-CID, and StackOverflow, respectively, suggesting that the optimal sampling strategy can be dataset-dependent.

\end{document}